\pdfoutput=1
\documentclass[11pt]{article}

\usepackage[preprint]{acl}

\usepackage{times}
\usepackage{latexsym}

\usepackage[T1]{fontenc}
\usepackage[utf8]{inputenc}

\usepackage{microtype}

\usepackage{inconsolata}

\usepackage{graphicx}

\usepackage{amsmath}
\usepackage[ruled]{algorithm2e}
\usepackage{enumitem}

\usepackage{rotating}
\usepackage{graphicx}
\usepackage{longtable}
\usepackage{pdflscape}

\usepackage{booktabs}
\usepackage{colortbl}
\usepackage{hyperref}
\usepackage{subcaption}
\usepackage{array}
\usepackage[normalem]{ulem}
\useunder{\uline}{\ul}{}
\usepackage{listings}
\lstdefinelanguage{json}{
    basicstyle=\ttfamily\small,
    showstringspaces=false,
    breaklines=true,
    backgroundcolor=\color{gray!10},
    stringstyle=\color{blue},
    keywordstyle=\color{magenta},
    commentstyle=\color{gray},
    morekeywords={true,false,null}
}

\definecolor{dark_green}{RGB}{22,120,22}

\title{Open Grounded Planning: Challenges and Benchmark Construction}

\author{
    Shiguang Guo\textsuperscript{1,3}\thanks{Equal contribution}\quad
    Ziliang Deng\textsuperscript{1,3}\footnotemark[1]\quad
    Hongyu Lin\textsuperscript{1}\thanks{Corresponding author}\quad
    Yaojie Lu\textsuperscript{1}\footnotemark[2]\\
    \textbf{Xianpei Han\textsuperscript{1,2,4}}\quad
    \textbf{Le Sun\textsuperscript{1,2,4}}\quad
    \\
    \textsuperscript{1}Chinese Information Processing Laboratory 
    \textsuperscript{2}State Key Laboratory of Computer Science\\ 
    Institute of Software, Chinese Academy of Sciences, Beijing, China \\
    \textsuperscript{3}University of Chinese Academy of Sciences, Beijing, China \\
    \textsuperscript{4}Key Laboratory of System Software, Chinese Academy of Sciences \\
  \texttt{\{guoshiguang2021,dengziliang2021,hongyu,luyaojie\}@iscas.ac.cn}\\
  \texttt{\{xianpei,sunle\}@iscas.ac.cn}}

\begin{document}
\maketitle
\begin{abstract}
The emergence of large language models (LLMs) has increasingly drawn attention to the use of LLMs for human-like planning. Existing work on LLM-based planning either focuses on leveraging the inherent language generation capabilities of LLMs to produce free-style plans or employs reinforcement learning approaches to learn decision-making for a limited set of actions within restricted environments. However, both approaches exhibit significant discrepancies between the open and executable requirements in real-world planning. In this paper, we propose a new planning task---open grounded planning. The primary objective of open grounded planning is to ask the model to generate an executable plan based on a variable action set, thereby ensuring the executability of the produced plan. To this end, we establish a benchmark for open grounded planning spanning a wide range of domains. Then we test current state-of-the-art LLMs along with five planning approaches, revealing that existing LLMs and methods still struggle to address the challenges posed by grounded planning in open domains. The outcomes of this paper define and establish a foundational dataset for open grounded planning, and shed light on the potential challenges and future directions of LLM-based planning. Our code and datasets are at \url{https://github.com/Shiguang-Guo/Open-Grounded-Planning}

\end{abstract}

\section{Introduction}

\begin{figure}[ht]
    \centering
    \includegraphics[scale=0.275]{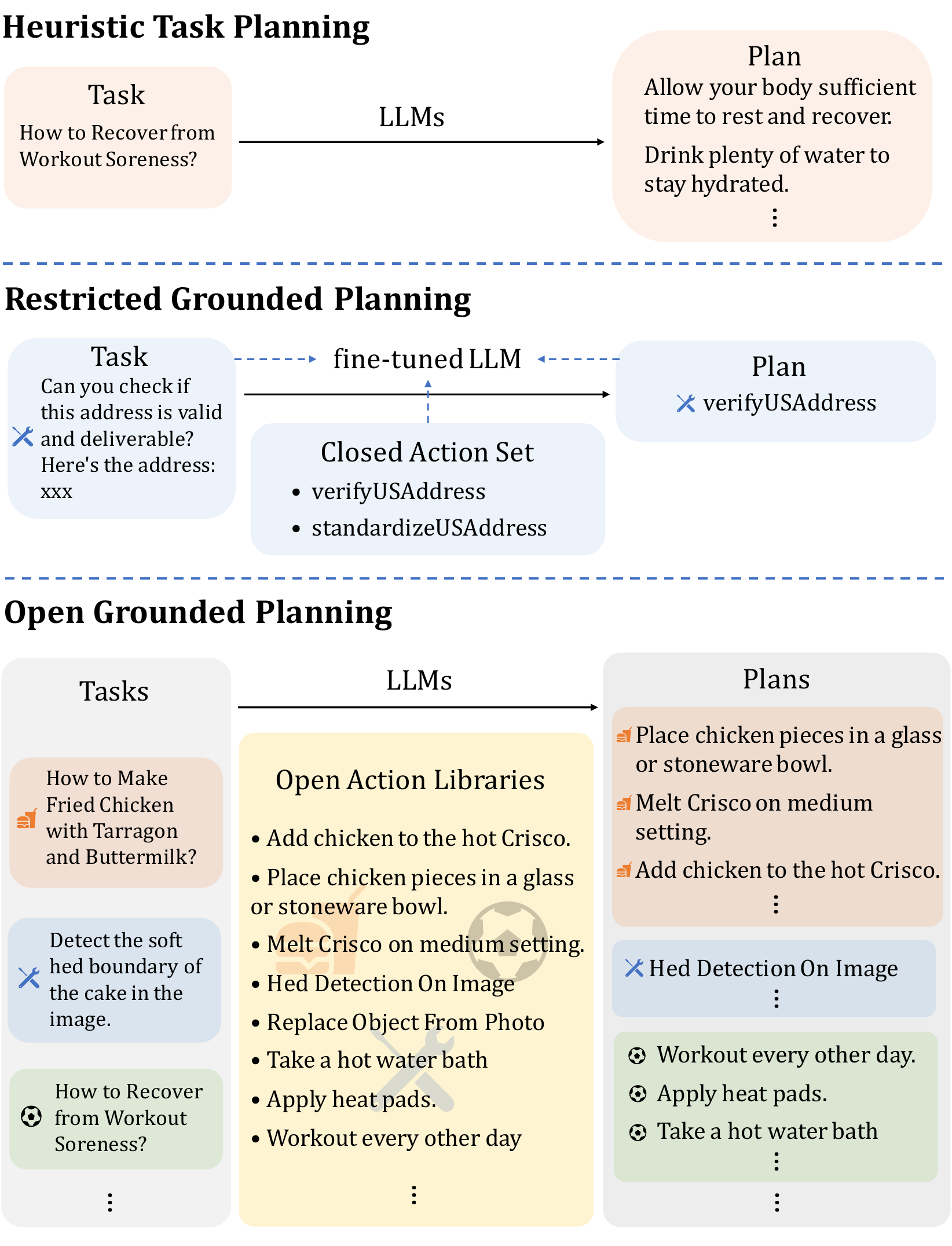}
    \caption{\textbf{Heuristic Task Planning}: Free and arbitrary planning. \textbf{Restricted Grounded Planning}: Domain-specific planning on small action sets, usually given in a context window. \textbf{Open Grounded Planning}: Planning on extensive action sets in various domains.}
    \label{fig:concept}
\end{figure}

Human life is filled with tasks of varying complexities, from simple activities like brewing coffee to more substantive pursuits such as learning new skills. By utilizing our understanding of the world, we can formulate plans for tasks and execute these steps in sequence. Although we can employ innumerable strategies and plans to achieve our objectives, the scenario is significantly more complex for artificial intelligence. Grounding plans to open action sets for tasks in open domains poses one of the challenges for AI. 

Some prior research has delved into the planning ability of Large Language Models (LLMs) and found that LLMs can engage in planning to some extent using their internal knowledge through common-sense reasoning \citep{zhao2023LargeLanguageModels, NEURIPS2020_1457c0d6}. However, these plans are often heuristic, coherent, and rational in natural language, yet possess a high degree of freedom and cannot serve as executable instructions for AI agents \citep{yao2023ReActSynergizingReasoninga,huang2022LanguageModelsZeroShot}. In other words, these plans are not grounded in an actionable space. In addressing the issue of grounded planning, various approaches have been explored in fields like robot controlling \citep{ahn2022CanNotSay, wang2023JARVIS1OpenWorldMultitask} and tool use \citep{qin2023ToolLLMFacilitatingLarge,liAPIBankComprehensiveBenchmark2023,tang2023ToolAlpacaGeneralizedTool}. Typically, model fine-tuning is applied for performance improvement in certain restricted scenarios \citep{song2023LLMPlannerFewShotGrounded, shen2023HuggingGPTSolvingAIa, yuan2023TaskLAMAProbingComplex}. However, these methods can only enable models to perform planning on a limited set of actions for specific domain tasks \citep{lin2023GroundedPlanningEmbodied, wu2023EmbodiedTaskPlanning,hao2023ReasoningLanguageModela}. As the task domain becomes broad, the action space becomes vast and open, these grounded planning methods appear too restricted to handle the steeply increased complexity \citep{wang2023JARVIS1OpenWorldMultitask}.

The capability to perform a wide range of actions and to devise viable, comprehensive plans by selecting suitable actions from an extensive action library for tasks in various domains epitomizes both a vision and a future trend in LLMs. Consequently, in this work, we introduce the concept of \textbf{\textit{Open Grounded Planning}} to advance the research on LLM planning across broad fields and a rich array of potential actions. We delineate the concept in two distinct dimensions:

\begin{itemize}
    \item \textbf{Grounded Planning:} LLM is required to compose plans utilizing only the actions available within the executable action sets.
    \item \textbf{Open Planning:} We aspire for the model to conduct planning within an extensive set of actions in an open domain that contains various task fields.
\end{itemize}

Moreover, we collect datasets from three major areas, including daily life, tool use, and robot sandbox scenarios. All collected datasets have been transformed into a uniform format, including task objectives, constraint conditions, golden steps, and candidate action sets. Building upon this foundation, we have developed a benchmark to assess the performance of various models and methods in the Open Grounded Planning task.

Besides, to address the open grounded planning challenges, we proposed a novel \textit{Retrieve and Rewrite} framework. The method utilizes the LLMs to generate an initial plan and iteratively rewrites this plan using actions retrieved based on the current planning situation.

We conducted comprehensive experiments on four commonly used methods and our \textit{Retrieve and Rewrite} method for current planning tasks using GPT-3.5, Vicuna-7B, and LLaMA-2-7B fine-tuned with a small amount of domain knowledge. The explored methods include retrieval-based methods and inference-based methods. We observed that fine-tuning contributes much to bridge the gap between smaller models and extremely large-scale language models by raising the instruction following and task understanding abilities. Various methods exhibited trade-offs regarding the executability and quality of generated plans. Generalizing ability from in-domain to out-of-domain planning tasks exists to a certain extent.

Generally speaking, our contributions are:

\begin{itemize}
    \item We proposed the concept of Open Grounded Planning. We envision future artificial intelligence systems being able to plan tasks within open domains, and having the ability to ground plans onto open executable action sets.
    \item We constructed a benchmark consisting of datasets from diverse domains for Open Grounded Planning and an automated evaluating procedure to assess the performance of different models and methods.
    \item We introduced the \textit{Retrieve and Rewrite} framework to address challenges in Open Grounded Planning tasks, and conducted comprehensive experiments on state-of-the-art models with various methods, and found that current models and methods still struggle with Open Grounded Planning tasks.
\end{itemize}

\section{Related Work}
\label{sec:related_work}
Large language models are trained on data containing extensive common knowledge and exhibit certain planning and common-sense reasoning abilities \citep{zhao2023LargeLanguageModels, NEURIPS2020_1457c0d6}. Prompting can be employed to guide large language models in generating plans for given tasks, and these plans often possess a high degree of freedom, making them challenging to execute in specific environments \citep{huang2022LanguageModelsZeroShot}. To facilitate the grounding of generated plans to an AI agent's executable action space, prior research has extensively explored grounded planning tasks \citep{lin2023GroundedPlanningEmbodied, wu2023EmbodiedTaskPlanning}. Some approaches opt for a global planning strategy based on the task, aiming to directly generate plans that can be grounded to the execution environment in a single step \citep{song2023LLMPlannerFewShotGrounded, shen2023HuggingGPTSolvingAIa, yuan2023TaskLAMAProbingComplex, wang2023DescribeExplainPlan}. Conversely, other methodologies employ iterative interactive approaches as the primary means of plan generation to adapt to changes in the environment and conditions \citep{ahn2022CanNotSay, wang2023JARVIS1OpenWorldMultitask}. However, these approaches often demonstrate limited effectiveness, completing constrained tasks with a finite set of actions within a singular domain.

In open-domain environments, the enormity of tasks and action sets poses significant challenges, making it increasingly difficult to bridge the gap between plans generated by large language models and the execution of real-world tasks. Therefore, in our work, we raise the challenge of open grounded planning and compile benchmark data from multiple domains ranging from everyday life to tool use and robot control scenarios, which consist of tens of thousands of tasks and actions. We also utilized our benchmark to assess the performance of mainstream proprietary models and open-source models with various planning methods on open grounded planning tasks.

\section{Open Grounded Planning Benchmark}
In this section, we initially present the task definition of Open Grounded Planning and its associated challenges. Subsequently, we introduce the Open Grounded Planning Benchmark, encompassing the construction of the dataset, evaluation metrics, and the methodology for automated assessment.

\subsection{Definition of Open Grounded Planning}
Specifically, for a given task objective $G$ stemming from any domain, along with conditional constraints $C$ (which may be absent for task without additional constraints), we aim to find a plan $P$ composed of a series of actions $\{s_i\}$, where each action $s_i$ is from an open action set $S$. In other words, the generated plan $P$ must be grounded onto the action set $S$ which is vast and extendable:
$$
P=(s_1,s_2,\cdots,s_n \space | \space G,C),s_i\in S,0\leq i\leq n
$$
where $n$ is the length of $P$. Table~\ref{tab:task_definition} shows the grounded planning process.

\begin{table}[htbp]
\centering

\resizebox{\linewidth}{!}{
\begin{tabular}{|l|}
\hline
\rowcolor[HTML]{C0D8CC} 
Task: \\
How to Activate the Dark Theme on YouTube \\
\rowcolor[HTML]{BDD7EE}
Method: \\
Using the YouTube App for Android \\
\hline
\rowcolor[HTML]{EFBCAE} 
Action Candidate Set: \\
* Close the Tool Options window. \\
* Double click the file. \\
* Do price forecasting. \\
* Click on the blue coloured YOUTUBE STUDIO BETA button. \\
* Open the YouTube app on your iPhone or iPad. \\
* Launch the YouTube app on your Android device. \\
* <other steps>... \\
\hline
\rowcolor[HTML]{F8E5C6} 
Steps: \\
1. Launch the YouTube app on your Android device. \\
2. Tap on your profile picture. \\
3. Tap on Settings. \\
4. Select the General option. \\
5. Tap on the grey switch, right across Dark theme text. \\
6. Enjoy YouTube in dark mode \\
\hline
\end{tabular}
}
\caption{An example of an Open Grounded Planning task. LLM needs to select appropriate actions from a complex and huge set of actions to generate a plan to complete the task.}
\label{tab:task_definition}
\end{table}

% \begin{table*}[htbp]
% \centering
% \begin{tabular}{ccccc}
% \hline
% \textbf{Category} & \textbf{Eval Set} & \textbf{} & \textbf{Subcategory} & \textbf{Eval Set} \\ \hline
% \multicolumn{2}{c}{WikiHow} &  & Education and Communications & 500 \\
% Computers and Electronics & 500 &  & Holidays and Traditions & 340 \\
% Relationships & 345 &  & Sports and Fitness & 500 \\
% Finance and Business & 500 &  & Work World & 402 \\
% Food and Entertaining & 500 &  & Personal Care and Style & 500 \\
% Home and Garden & 500 &  & Pets and Animals & 272 \\ \cline{4-5} 
% Travel & 375 &  & \multicolumn{2}{c}{Tools} \\
% Hobbies and Crafts & 500 &  & ToolAlpaca & 201 \\
% Family Life & 406 &  & API-Bank & 263 \\
% Health & 500 &  & GPT4Tools & 500 \\ \cline{4-5} 
% Arts and Entertainment & 500 &  & \multicolumn{2}{c}{Robot} \\
% Youth & 443 &  & VirtualHome & 500 \\
% Philosophy and Religion & 407 &  & SayCan & 164 \\
% Cars and Other Vehicles & 500 &  &  &  \\ \hline
% \end{tabular}
% \caption{statistics of evaluation set}
% \label{tab:evaluation-set}
% \end{table*}

% Please add the following required packages to your document preamble:
% \usepackage{booktabs}

\begin{table*}[htbp]
\centering
\resizebox{\linewidth}{!}{
\begin{tabular}{@{}ccccccccc@{}}
\toprule
\textbf{Category} & \textbf{Eval-Set} & \textbf{Full-Set} & \textbf{Actions} & \textbf{} & \textbf{Category} & \textbf{Eval-Set} & \textbf{Full-Set} & \textbf{Actions} \\ \midrule
% \rowcolor[HTML]{EFEFEF}
\multicolumn{9}{c}{wikiHow\citep{zhang2020ReasoningGoalsSteps}} \\ \midrule
Arts and Entertainment & 500 & 4104 & 26222 &  & Home and Garden & 500 & 6916 & 39872 \\
Cars and Other Vehicles & 500 & 1685 & 10929 &  & Personal Care and Style & 500 & 3888 & 20786 \\
Computers and Electronics & 500 & 12801 & 75186 &  & Pets and Animals & 500 & 2282 & 11056 \\
Education and Communications & 500 & 5485 & 29856 &  & Philosophy and Religion & 500 & 748 & 5000 \\
Family Life & 500 & 1532 & 8634 &  & Relationships & 500 & 1683 & 8609 \\
Finance and Business & 500 & 4376 & 24746 &  & Sports and Fitness & 500 & 1898 & 10916 \\
Food and Entertaining & 500 & 9493 & 58585 &  & Travel & 500 & 852 & 5433 \\
Health & 500 & 7918 & 37364 &  & Work World & 500 & 1088 & 6618 \\
Hobbies and Crafts & 500 & 7095 & 47168 &  & Youth & 500 & 1477 & 8389 \\
Holidays and Traditions & 500 & 904 & 5658 &  &  &  &  &  \\ \cmidrule(r){1-4} \cmidrule(l){6-9} 
% \rowcolor[HTML]{EFEFEF}
\multicolumn{4}{c}{Tools} &  & \multicolumn{4}{c}{Robot} \\ \cmidrule(r){1-4} \cmidrule(l){6-9} 
APIBank\citep{liAPIBankComprehensiveBenchmark2023} & 263 & 263 & 101 &  & SayCan\citep{ahn2022CanNotSay} & 164 & 164 & 97 \\
GPT4Tools\citep{yangGPT4ToolsTeachingLarge2023} & 500 & 1750 & 32 &  & VitualHome\cite{huang2022LanguageModelsZeroShot} & 500 & 5088 & 47522 \\
ToolAlpaca\citep{tang2023ToolAlpacaGeneralizedTool} & 201 & 201 & 89 &  &  &  &  &  \\ \bottomrule
\end{tabular}
}
\caption{Statistics of the Open Grounded Planning benchmark, marking the quantitative attributes of the in-domain and out-of-domain datasets.}
\label{tab:evaluation-set}
\end{table*}

As discussed in Section \ref{sec:related_work}, many explorations into LLM planning are focusing on heuristic planning in which the generated plans cannot be directly used as instructions for downstream control mechanisms, in other words, they are not "grounded". Some previous studies have demonstrated that LLMs can undertake grounded planning tasks in certain fields. However, these applications have often been limited to constrained scenarios and task domains. As the richness of the task domains and actions increases, the model's planning proficiency tends to diminish. LLMs still face challenges in executing grounded planning across open domains, which encompass a wide array of tasks and actions from diverse fields.

\subsection{Dataset Construction}
\label{sec:odp_benchmark}
LLM's planning capabilities have a variety of application scenarios. We refer to many other works and summarize the three main application areas including daily life, tool usage, and robots. To balance the proportions of data across different categories, we retain a maximum of 500 tasks for each category, forming our evaluation set. All actions related to the original tasks are preserved in the action library as candidate actions. 

We split the dataset into two parts. We employ the daily life dataset wikiHow to evaluate the in-domain grounded planning capabilities because this dataset covers a very wide range and the action set for selection is more complex. Additionally, we utilize datasets related to tool use and robots to evaluate the generalization of various models and methods for out-of-domain grounded planning. The statistical information of the evaluation set can be found in Table~\ref{tab:evaluation-set}. We also provide the more detailed data processing procedure in Appendix~\ref{sec:appendix-dataset}.

\subsubsection{In-Domain Datasets}

\paragraph{Wikihow} Wikihow is an extensive collection of guides and tutorials, encompassing topics ranging from everyday life skills to more complex subjects\footnote{\url{https://www.wikihow.com/}}. Each guide on WikiHow is presented in a step-by-step manner, making it easy to understand and follow. We gathered the original corpus of WikiHow by referencing \citet{zhang2020ReasoningGoalsSteps}. For each article, we retained only the tasks, methods (if any), and headlines. We eliminated sections containing multiple "parts" as they introduced additional hierarchy. By directly utilizing the original categorization within the WikiHow corpus, we ultimately identified 19 categories, with a total of more than 76,000 tasks. The action libraries for each category are derived from the collective actions of all tasks within the same category, with an average size exceeding 20,000.

\subsubsection{Out-of-Domain Datasets}

\paragraph{Tools} Previous studies have demonstrated the capability of LLMs to utilize tools to accomplish tasks. Effective planning is crucial for tool use, especially when the candidate toolset is extensive. We have collected open-source data relevant to tool usage by LLMs, including contributions from ToolAlpaca \citep{tang2023ToolAlpacaGeneralizedTool}, API-Bank \citep{liAPIBankComprehensiveBenchmark2023}, and GPT4Tools \citep{yangGPT4ToolsTeachingLarge2023}. These datasets encompass various types of tools and provide standard tool invocation sequences to complete the tasks as well. To maintain consistency with other datasets, we only retain the API names and their corresponding description, while ignoring the parameters. 

\paragraph{Robot} There exists some research related to grounded planning in robotics\citep{yoshida2023TextMotionGrounding,brohan2023RT1RoboticsTransformer,ahn2022CanNotSay}, but there is still a lot of room for development. We have converted datasets proposed in VirtualHome \citep{huang2022LanguageModelsZeroShot} and SayCan \citep{ahn2022CanNotSay} and merged all executable actions as a candidate action set\footnote{There seems to be some mismatch in the dataset provided by saycan, we fix it manually. A more detailed procedure is provided in Appendix~\ref{sec:appendix-dataset}.}. It is important to note that complete robot processing involves multiple stages, including visual information processing and action execution. Our dataset, however, only focuses on the planning generation.

% \begin{figure*}[htbp]
%     \centering
%     \includegraphics[scale=0.49]{sections/figures/3methods.pdf}
%     \caption{3 Methods}
%     \label{fig:3methods}
% \end{figure*}

\subsection{Evaluation}
\label{sec:evaluation}

\subsubsection{Plan Quality Assessment}
\label{sec:Multi-dimensional evaluation}

In all the datasets we collected, every task has a corresponding golden plan which is provided by the original datasets and presents one of the possible ways to handle the task. Since the solutions to the tasks in our benchmark could be quite diverse, especially when it involves thousands of candidate operations to form a planned execution path, it is unfair to directly judge whether the generated plans match exactly with the golden plans. Instead, we make the golden plan a reference and compare the plan generated by the model with it from multiple perspectives to judge which plan is better. The specific evaluation criteria are as follows: 

\textbf{Completeness}: Examine whether the plan is comprehensive, with a focus on the coherence and logic between steps, and the avoidance of arbitrarily introduced conditions and missing steps. 

\textbf{Feasibility}: Assess the practicality of the plan, considering whether each step can be implemented, whether the plan aligns with common sense, adheres to human ethical standards, and avoids excessive redundant steps. 

\textbf{Relevance to the Task}: Evaluate the relevance of the plan to the given task, considering the utilization of the provided task conditions and whether it achieves the goal.

We use ChatGPT to evaluate, which is a widely used evaluation method in previous similar work. During one evaluation, the target task to be solved is delivered to the evaluator along with the model generated plan and the golden plan. Then we prompt ChatGPT to list how good the two plans are regarding the evaluation criteria, and ask it to elect a better plan based on its analyses.

Due to many reported issues with ChatGPT as an evaluator, such as position bias, length preference, and style partiality \citep{koo2023BenchmarkingCognitiveBiases,wu2023StyleSubstanceEvaluation,zheng2023LargeLanguageModels}, we employ various methods to mitigate those biases. We swap the order of the two plans and average the scores to eliminate positional bias. Additionally, we prompt ChatGPT to penalize the score for redundant steps to reduce length preference. We sampled a small dataset for manual evaluation and verified the plausibility of ChatGPT's automatic evaluation. For detailed analysis, please refer to appendix~\ref{sec:appendix-evaluation}.

\subsubsection{Metrics}
\label{sec:metrics}

To more intuitively compare the performance of various models and methods on open domain planning datasets, we define the following metrics to quantify their performance.

\textbf{Executability} is the proportion of executable cases. Executable cases are actions in the plan that all exist within the given action library.

\textbf{Quality} of the executable plans is evaluated from the dimensions in section \ref{sec:Multi-dimensional evaluation}. We define win rate as the average of the outcomes of two comparisons involving position swaps. Intuitively, quality assesses how complete the generated plan is.

\textbf{Overall Pass Rate} is the proportion of all generated plans that can be executed while also completing the task. Considering both executability and quality, we choose pass rate as the final evaluation metric to evaluate the overall performance of the model in the entire process. The pass rate is the product of executability and quality.
\begin{align*}
    \text{Executability} &= \frac{\#\text{executable cases}}{\# \text{all cases}}\\
    \text{Quality} &= \frac{\#\text{win cases}}{\# \text{executable cases}}\\
    \text{Pass Rate} &= \text{Executability} \times \text{Quality}
    % \text{Out-of-Actions} &= \frac{\#\text{ hallucination steps}}{\# \text{all cases}-\#\text{bad cases}}
\end{align*}

\section{Methods}
\label{sec:method}
\begin{figure*}[htbp]
    \centering
    \includegraphics[width=\linewidth]{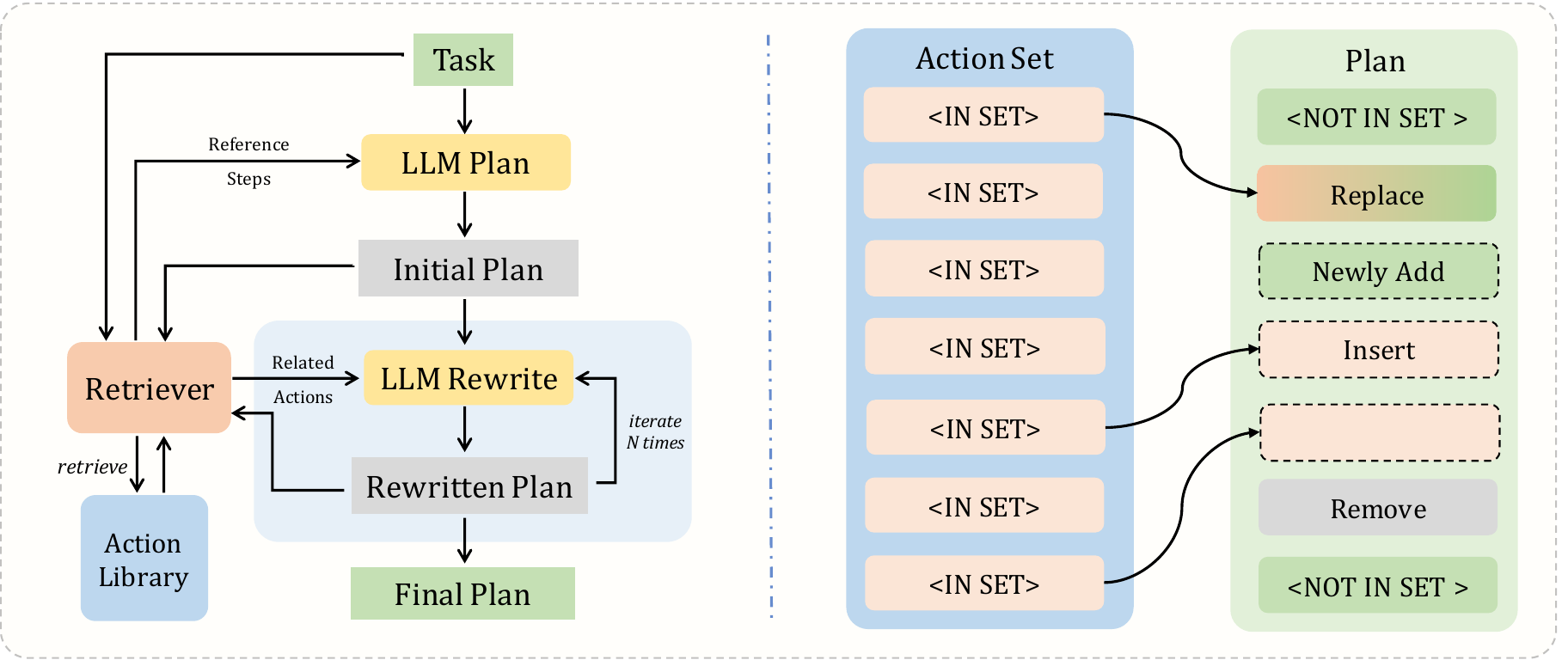}
    \caption{\textbf{Left}: The Retrieve and Rewrite framework. \textbf{Right}: An illustration of different rewriting operations.}
    \label{fig:rewrite}
\end{figure*}

In order to assess the performance of current main-stream models on the Open Grounded Planning task, we endeavored to employ five distinct methods including \textit{Retrieve and Rewrite}, a new framework we proposed, to address this challenge. 

\paragraph{Task-Retrieve:} We first adopt a simple and intuitive approach by using the task name as the query for action retrieval. Given a task $T$ and an action set $A$, we retrieve the relevant action list $A^{\prime} = \text{retrieve}(T, A)$. LLM selects and orders appropriate actions from this list to generate plan $P=\text{select\&sort}(T, A^{\prime})$.

\paragraph{Plan-Retrieve:} Simply searching by task name makes it difficult to recall important steps that are not directly related to the task name. We try to specify the query to improve. We first force LLM to generate an initial plan $P_0=\left\{s_1, s_2,\cdots, s_n\right\}$ for the task, then retrieve related actions based on the generated plan to get action list $A^{\prime} = \text{retrieve}(P_0, A)$. Finally, we perform similar selection and rearrangement as in task-retrieve. We also provide the initial plan for LLM to refer to generate the final plan $P=\text{select\&sort}^{\prime}(T, A^{\prime}, P_0)$.

\paragraph{Step-wise Select:} In addition to retrieval and rearrangement, we also try the step-wise selection method. We adopt a method similar to ReAct \citep{yao2023ReActSynergizingReasoninga} to our tasks. Each time, LLM generates a possible next step $s_p$ for the given task $T$, Then we obtain a candidate action list $A'=\{a_1,a_2,...\} \subset A$ by retrieving actions based on the generated steps. LLM selects one of the retrieved results as the next step, which means the target plan $P$ is generated step by step, i.e., $P=\{s_1, s_2,...\}\cup\{a_j\}$ where $a_j \in A^{\prime}$. Selection iteration repeats until 1) LLM outputs \textit{None} when generating the possible next step, 2) LLM refuses to select from the candidate list, and 3) the maximum number of iterations is reached.

\paragraph{DFS:} The \textit{Step-wise Selecting} method suffers from small searching space. ToolLLM \citep{qin2023ToolLLMFacilitatingLarge} proposed a DFS-based method to improve. We implement this by simply extending the \textit{Step-wise Selecting} method. During the procedure of model selecting the next step, we allow LLM to abandon the selection if it thinks there exists no suitable choice in the retrieved candidate actions to be the next step of the current plan. In this case, we perform a backtracking. 

\paragraph{Retrieve and Rewrite:} We realize that methods based on retrieval and rearrangement can consider the overall plan, but may not obtain the optimal choice for each step. The step-wise selection approach enables adjustments during the generation process but does not take into account the plan as a whole. We combine the advantages of both methods and propose a new method named \textit{Retrieve and Rewrite}.Figure~\ref{fig:rewrite} illustrates its framework and different rewriting operations.

LLM is first asked to generate an initial plan $P_0=\{\underline{s}_1, \underline{s}_2, ..., \underline{s}_{n_{0}}\}$ based on the relevant steps $A_0$ with the given task $T$. Different from the \textit{Task-Retrieve} method, $P_0$ does not have to be composed of the exact steps in $A_0$. We mark steps not in action set with underline. We perform several iterations to use steps from the action set to rewrite $P_0$. For iteration $i\ge1$, we choose some of the steps not in action set for retrieval to retrieve relevant actions $A_{i}=\{a_{i1}, a_{i2}, \cdots, a_{im_{i}}\}$, where $m_i$ is the length of candidate action list of iteration $i$. LLM is allowed to perform various operations when rewriting $P_{i-1}$, including adding, deleting, and modifying arbitrarily. We only need to ensure using actions \textit{in} the action set to replace as much as possible those \textit{not in} the action set. The plan after rewriting might be $P_{i}=\left\{s_1,\underline{s}_2,s_3,\cdots,\underline{s}_{n_i}\right\}$, where $n_i$ is the new length of current plan $P_{i}$. Similarly, iteration stops until all actions are in the action set or iteration reaches the maximum number.

\section{Experiment}
We systematically assess the capabilities of various LLMs and methods in the Open Grounded Planning task by selecting and comparing mainstream proprietary models, open-source models, and open-source models fine-tuned with a small amount of domain-specific data. For each model, we examine the performance of different methods in Section \ref{sec:method}. We test the models' abilities in Open Grounded Planning on both in-domain, the wikiHow dataset, and out-of-domain, the tools and robot datasets.

\subsection{Experiment Settings}

\begin{table*}[ht]
\resizebox{\linewidth}{!}{
\begin{tabular}{lccc|ccccc}
\toprule
 & \multicolumn{3}{c}{\textbf{In-Domain (wikiHow)}} & \multicolumn{5}{c}{\textbf{Out-of-Domain (Tools, Robot)}} \\
\cmidrule(r){2-4} \cmidrule(l){5-9} 
 & \multicolumn{3}{c}{Average of All Types} & APIBank & GPT4Tools & ToolAlpaca & VirtualHome & SayCan \\
 \cmidrule(r){2-4} \cmidrule(l){5-7} \cmidrule(l){8-9}
Method & \multicolumn{1}{c}{Executability(\%)} & \multicolumn{1}{c}{Quality(\%)} & \textbf{Pass Rate(\%)} & \textbf{Pass Rate(\%)} & \textbf{Pass Rate(\%)} & \textbf{Pass Rate(\%)} & \textbf{Pass Rate(\%)} & \textbf{Pass Rate(\%)} \\ \midrule
\rowcolor[HTML]{EFEFEF}
\multicolumn{9}{c}{\cellcolor[HTML]{EFEFEF}Vicuna-7B-v1.5-16k} \\ \midrule
Task-Retrieve & {\color{gray}89.60} & {\color{gray}27.87} & 24.97 & 36.50 & 16.30 & 19.90 & 12.40 & 14.94 \\
Plan-Retrieve & {\color{gray}67.77} & {\color{gray}42.41} & 28.74 & 26.05 & 15.10 & 9.70 & 11.90 & 12.20 \\
Step-wise Select & {\color{gray}73.17} & {\color{gray}12.34} & 9.03 & 20.72 & 15.80 & 10.20 & 7.10 & 1.83 \\
DFS & {\color{gray}97.92} & {\color{gray}7.18} & 7.03 & 26.81 & 20.50 & 16.17 & 6.80 & 2.44 \\
Retrieve and Rewrite & {\color{gray}80.75} & {\color{gray}34.88} & 28.17 & 45.44 & 22.30 & 23.63 & 16.00 & 12.50 \\ \midrule
\rowcolor[HTML]{EFEFEF}
\multicolumn{9}{c}{\cellcolor[HTML]{EFEFEF}GPT-3.5} \\ \midrule
% Task-Retrieve & {\color{gray}95.99} & {\color{gray}44.43} & 42.65 & 7.22 & 23.20 & 5.97 & 26.50 & 29.27 \\
Task-Retrieve & {\color{gray}95.99} & {\color{gray}44.43} & 42.65 & 37.26 & 23.20 & 14.93 & 26.50 & 29.27 \\
Plan-Retrieve & {\color{gray}69.46} & {\color{gray}60.15} & 41.78 & 30.61 & 32.30 & 11.94 & 37.60 & \textbf{44.82} \\
Step-wise Select & {\color{gray}93.44} & {\color{gray}21.21} & 19.82 & 32.32 & 23.50 & 16.67 & 30.60 & 26.83 \\
DFS & {\color{gray}98.84} & {\color{gray}50.76} & 50.17 & 35.55 & 25.00 & 19.90 & 32.90 & 11.28 \\
Retrieve and Rewrite & {\color{gray}92.98} & {\color{gray}58.72} & 54.60 & 43.73 & 26.60 & 28.86 & \textbf{47.00} & 41.16 \\ \midrule
\rowcolor[HTML]{EFEFEF}
\multicolumn{9}{c}{\cellcolor[HTML]{EFEFEF}LLaMA-2-7B(SFT)} \\ \midrule
Task-Retrieve & {\color{gray}99.40} & {\color{gray}47.66} & 47.37 & \textbf{58.75} & 26.50 & \textbf{49.00} & 37.50 & 37.50 \\
Plan-Retrieve & {\color{gray}99.13} & {\color{gray}58.21} & {\ul 57.70} & {\ul 47.72} & {\ul 35.00} & 36.07 & 42.70 & 30.79 \\
Step-wise Select & {\color{gray}99.82} & {\color{gray}24.26} & 24.22 & 36.12 & 4.20 & 21.89 & 34.10 & 31.71 \\
DFS & {\color{gray}99.09} & {\color{gray}53.53} & 53.04 & 11.40 & 0.48 & 2.79 & 35.64 & 14.96 \\
Retrieve and Rewrite & {\color{gray}98.26} & {\color{gray}61.58} & \textbf{60.51} & 45.42 & \textbf{43.70} & {\ul 45.02} & {\ul 46.80} & {\ul 42.68} \\ \bottomrule
\end{tabular}
}
\caption{The average performance of models and methods on in-domain and out-of-domain datasets. The final metric is \textbf{Pass Rate}. The best performance score of each dataset is highlighted with bold, while the second-best underlined.}
\label{tab:whole}
\end{table*}

We chose the proprietary model GPT-3.5\footnote{We use \textit{gpt-3.5-turbo-1106} for our experiments.} and the open-source model Vicuna-7B-v1.5-16k \citep{zheng2023JudgingLLMasaJudgeMTBencha} for experiments. In addition to this, we fine-tuned Llama-2-7B \citep{touvron2023LlamaOpenFoundationa} to check the performance of the SFT model. We believe these three models can represent the capabilities of current mainstream models.

We select 200 tasks from each subcategory below wikiHow as the training set. For each setting, we perform the inference process of GPT-3.5 on it and select those with high quality for training. We mixed it with the Alpaca dataset \citep{alpaca} and fine-tuned the model with 3 epochs to improve the generalization ability. We use OpenAI's \textit{text-embedding-ada-002} to generate embedding for each step in all settings. Additional implementation details are in Appendix~\ref{sec:appendix-details} and all prompts are in Appendix~\ref{sec:appendix-prompt}.

\subsection{In-Domain Results}
\label{subsec:main_results}
We measure the performance of each model and method using metrics in Section~\ref{sec:metrics}. We report the average performance on all wikiHow datasets. Results on in-domain datasets are presented in the left part of Table~\ref{tab:whole}, from which we can derive:

\textbf{SFT model achieves the best performance} Compared to Vicuna and GPT-3.5, the SFT model surpasses them in all methods. The trained model can improve the executability of all methods to close to 100\% and maintain high quality at the same time.

\textbf{Different methods have different focuses} Although the initial plan of the Plan-retrieve method may cause interference, it can generate a better plan than Task-retrieve. Compared with the restricted search space in the step-wise selection, DFS usually achieves a higher executability and has better quality. Besides, we find pre-planning grants the final plans higher quality. The Retrieve and Rewrite method we proposed surpasses Step-wise select and DFS in terms of rationality and completeness of the final plans due to our pre-planning approach and the subsequent rewriting.
\subsection{Out-of-Domain Results}

The experiment results on the out-of-domain datasets are presented in the right part of Table~\ref{tab:whole}. Due to space constraints, we provide more detailed OOD dataset results in Appendix~\ref{sec:appendix-ood}. Apart from the DFS method, LLaMA-2-SFT outperforms GPT-3.5 on most datasets using other methods. This suggests that fine-tuning is beneficial for the model to generalize from in-domain grounded planning to out-of-domain scenarios, primarily by enhancing the model's ability to follow instructions, thereby improving task completion. In the tools dataset, despite our extensive prompt engineering, GPT-3.5 prefers to generate complete steps rather than the provided API names, which results in poor performance.

We observe the performance degradation in SFT model with DFS and find that the average length of the plans generated with DFS on tools dataset is less than 1, while with other methods ranges from 1 to 2, which is also closer to the true length of datasets. We observe extensive backtracking in the generated results,which could be attributed to the higher proportion of rejection in the training data, leading the SFT model to exhibit reduced confidence in its generation. Consequently, this makes LLM more inclined not to select the retrieved results as the next step when facing OOD data.

Among various methods, Rewrite method consistently maintains strong performance. However, although it performs well in-domain, the generalization performance is sometimes not as good as simple retrieval and rearrangement.

\subsection{Error Analysis}
\begin{table}[htbp]
\centering
\resizebox{\linewidth}{!}{
\begin{tabular}{p{3cm}p{9cm}}
\hline
\multicolumn{1}{c}{Task} & \multicolumn{1}{c}{How to Build a Small Chick Brooder} \\ \hline
\multicolumn{1}{c}{Method} & \multicolumn{1}{c}{Setting up the Brooder} \\ \hline
\textbf{Plan-Retrieve}
 & 
\begin{minipage}[t]{9cm}
Set up a brooder.\newline 
Use absorbent bedding.\newline 
Provide a warm place.\newline 
\textcolor{red}{Place any heating apparatus needed for the species that is going to be kept.}\newline 
Monitor the temperature.\newline 
Set up an attractive feeder.\newline 
\textcolor{red}{Choose an appropriate hiding spot.}\newline 
\textcolor{red}{Ensure that there is proper ventilation in the stable.}
\end{minipage}
\vspace{0.2mm}\\ 
\hline
\textbf{DFS}
 & 
\begin{minipage}[t]{9cm}
Set up a brooder.\newline
Point a heat lamp on one side of your brooder to keep the chicks warm.\newline
Move your chicks to their brooder.\newline
\textcolor{red}{Move the chicks to a brooder.}\newline
Provide food and water for your chicks.\newline
Move the chicks to a brooder.\newline
\textcolor{red}{Set up a brooder.}\newline
\textcolor{red}{Move the chicks to a brooder.}
\end{minipage}
\vspace{0.2mm}\\ 
\hline
\textbf{Golden plan} & 
\begin{minipage}[t]{9cm}
Fill the bottom of your brooder with paper towels and pine shavings.\newline 
Point a heat lamp on one side of your brooder to keep the chicks warm.\newline 
Provide food and water for your chicks.\newline 
Put a branch in the brooder to create a roost.
\end{minipage}
\vspace{0.2mm}\\ 
\hline
\textbf{Rewrite} & 
\begin{minipage}[t]{9cm}
\textcolor{dark_green}{Position the cage away from drafts and direct sunlight.}\newline 
Set up a brooder.\newline 
Fill the bottom of your brooder with paper towels and pine shavings.\newline 
Put a branch in the brooder to create a roost.\newline 
Point a heat lamp on one side of your brooder to keep the chicks warm.\newline 
Move your chicks to their brooder.
\end{minipage} \\ 
\hline
\end{tabular}
}
\caption{\textcolor{red}{Red text} indicates steps that are irrelevant to the task and redundant steps, while \textcolor{dark_green}{green text} denotes supplementary additional information.}
\label{tab:case_study}
\end{table}
We mainly perform analysis on GPT-3.5 on wikiHow to fairly compare various methods. We analyze from the perspectives of executability and generation quality. As executability shown in table~\ref{tab:whole}, since the output format of the first four methods is usually a sentence or a list, inexecutable plans all come from the hallucination of LLM, which means LLM generate content beyond the given set. However, the output format of the Rewrite method is more complex. We observe that the non-executable plans contains 11.84\% format parsing errors. The small format error ratio proves that LLMs have strong instruction following ability, but they still face serious hallucination problem.

To compare the generation quality of different methods, we present cases from the Plan-retrieve, DFS, Rewrite and golden plan from wikiHow in table~\ref{tab:case_study}. The plans generated by Plan-retrieve might generate steps that are correct in meaning but irrelevant in detail to the task. This is because the retrieved actions are not always relevant to the task, and LLM, given a one-time, limited selection, may be forced to choose these steps to make the plan complete. Besides being incomplete or irrelevant to the task, DFS also suffers from duplicate steps, despite being provided with previously selected steps. We find that 19.32\% of plans generated by DFS contain repetitions of two or more times. Meanwhile, this step by step generation is also less complete, for example, missing information on the soft bedding material. In contrast, the Rewrite method, through iteration and global consideration, can generate more complete plans. Additionally, with a large pool of candidate steps,  LLM can even find supplementary information for the task.

\subsection{Retrieval Amount Influence}
\begin{figure}[ht]
    \centering
    \includegraphics[width=0.85\linewidth]{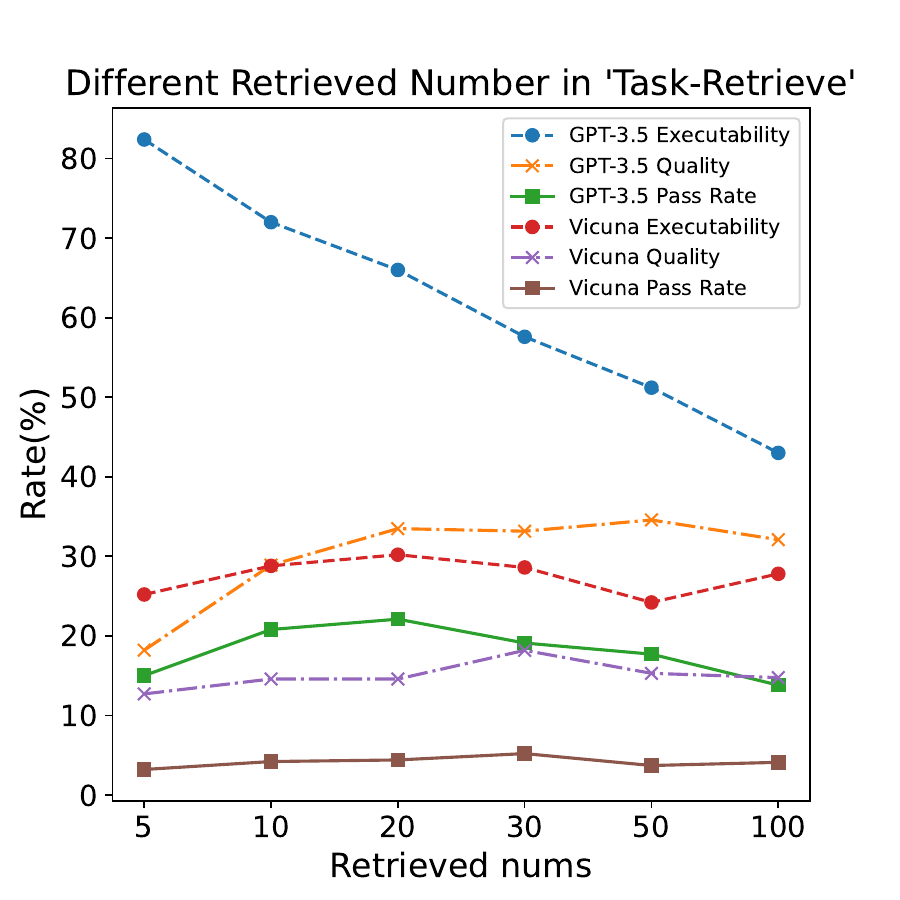}
    \caption{Influence of the number of retrieved actions on the performance of GPT-3.5 and Vicuna conducting Task-Retrieve method.}
    \label{fig:retrieve_num}
\end{figure}

We find that the number of retrieved actions directly affects the execution rate and quality of the generated plan. We check the result on \textit{wikiHow-Computers and Electronics} for a setup using GPT-3.5 and Vicuna for the task-retrieve method\footnote{Different from the main experiment, here we only check the output format.}. Figure~\ref{fig:retrieve_num} shows how "Executability", "Quality" and "Pass Rate" change when different numbers of steps are recalled for plan generation. The two solid lines representing "Pass Rate" demonstrate a trend of initial increase followed by a decrease. For GPT-3.5, as the number of recalled actions increases, the proportion of generated plans that conform to the rules decreases, while the quality of plans lifted. Pass Rate achieves the best performance when recalling 20 actions. Interestingly, we observed that when recalling 5 actions there are still parts of the plan that are not executable. We found that this is because the recalls are too little for LLMs to complete the task using the recall steps. But GPT-3.5 tends to generate complete plans, so steps for supplementing and connecting context are generated. We continue to discuss the results in other settings in Appendix \ref{sec:appendix-retrieved}.

\section{Conclusion}

In this study, we introduced \textit{Open Grounded Planning} and developed a benchmark comprising datasets from various domains with vast action sets. Extensive experiments revealed significant limitations in the performance of current models and methods in generating grounded plans for these sets. Furthermore, we observed a pronounced challenge in enabling these models and methods to generalize from in-domain scenarios to out-of-domain datasets. Compared to four other methods, our "Retrieve and Rewrite" approach demonstrates a partial resolution to the challenges inherent in open grounded planning. Our work highlights the need for enhancing the capability of models and methods for expansive planning domains and improving the executability and quality of grounded plans, laying a foundation for future research.

\section*{Limitation}
Our current implementation relies on a two-stage approach of retrieval and generation. We anticipate that an optimized retriever tailored for this task will achieve better performance. Additionally, our current dataset only explores how to ground tasks into a given set of actions. While this is sufficient for many applications, a more challenging extension would be to introduce parameters for each action, meaning the use of a collection of action functions instead of a collection of actions.

Our current method for evaluating the content of the plan involves using ChatGPT as the evaluator, which is to guide ChatGPT to compare and judge the plans through prompting. This method inevitably brings bias and hallucination into the evaluation results. Although we have employed various methods to alleviate their impact and have sampled some cases for manual evaluation to prove the effectiveness of our evaluation method, bias and hallucination persist. In future work, we may introduce more diverse objective evaluation metrics and ways to reduce bias and hallucination.
\section*{Acknowledgements}
We sincerely thank all anonymous reviewers for their insightful comments and valuable suggestions. This research work is supported by the National Natural Science Foundation of China under Grants no. 62106251, no. 62122077 and no. 62306303, and the Basic Research Program of ISCAS, Grant No. ISCAS-JCZD-202303.

\bibliography{reference/anthology,reference/LLM,reference/planning}

\newpage

\appendix
\section{Dataset Details}
\label{sec:appendix-dataset}
For the datasets of WikiHow, Tools, and Robot, although they have different original formats, they can all be transformed into a uniform format with three parts: task name, method, and steps. For WikiHow, each life guide (task) contains a title and several steps, with each step including a headline and a detailed explanation of the step. We retain only the headline part as the step to complete the task. Some guides may have different methods to achieve this. For example, one can create a poster by either hand drawing or using paper cutting. We randomly select one of these methods as the method to accomplish the task. The final example for WikiHow is shown in Table~\ref{tab:task_definition}.

For Tools scenarios, we extract all the API calls under each task as steps. To maintain consistency with other tasks, we only retain the API name and description without including API parameters, as this would require additional training, and many works have already explored this kind of capability\cite{qin2023ToolLLMFacilitatingLarge}. We will also set the default method to some tasks to make the results they generate more consistent with requirements. Here is an example from GPT4Tools. Given a task of \textit{Generate a real image of xxx from the sketch image} and a training instance for API call, we will transform it into a process of \text{Sketch Detection On Image} -> \textit{Generate Image Condition On Sketch Image}.

\lstset{language=json}
\begin{lstlisting}
{
    "title": "Generate a real image of a cat sitting on a table next to a bowl from the sketch image",
    "method": "One or two steps are usually enough to complete the task, and there are only a few cases where more may be required.",
    "steps": [
        "Sketch Detection On Image DESCRIPTION: useful when you want to generate a scribble of the image. like: generate a scribble of this image, or generate a sketch from this image, detect the sketch from this image. ",
        "Generate Image Condition On Sketch Image DESCRIPTION: useful when you want to generate a new real image from both the user description and a scribble image or a sketch image. "
    ]
}
\end{lstlisting}

In the Robot scenario, we can also obtain tasks, methods, and corresponding steps in the same way. As described in the footnotes, the SayCan repository provides two files: one providing tasks and environment states, and the other providing tasks and corresponding plans, but they do not completely match. Since the tasks in this dataset have similar processes, such as instructing the robot to reach a certain place or pick up an item, we filled in the inconsistent parts based on this style. We also provide an example here.

\lstset{language=json}
\begin{lstlisting}
{
    "title": "I'd like a clear soda.",
    "method": "As a robot with only one gripper, you are surrounded by a far counter, a near counter, a table, and a trash can. You are located near the table. You can only perform one action at a time, such as moving or picking up and putting down. Environmental status:water on table, 7up on table",
    "steps": [
        "find a 7up",
        "pick up the 7up",
        "bring it to you",
        "put down the 7up"
    ]
}
\end{lstlisting}

\section{Implementation Details}
\label{sec:appendix-details}
We use FastChat \citep{zheng2023JudgingLLMasaJudgeMTBencha} for training, and the training parameters are consistent with vicuna-7B. For all generation steps, we perform generation with temperature = 1.0. We perform up to five retries per generation to avoid formatting errors. We also performed a rule review on the output of LLM to obtain the best performance of LLM in the open grounding planning task. For example, if LLM is required to choose one of several options as the next step, and it does not output a sentence that meets the requirements, we will regard this output as a failure and regenerate it. This retry will count towards the five retries above. For tools datasets, our output combines API name and API description, and the input format is "\{api name\} DESCRIPTION: \{API description\}". Since we only care about choosing the correct step, we accept both API name and "\{API name\} DESCRIPTION: \{anything\}" as input.

We simply use OpenAI's \textit{text-embedding-ada-002} for embedding generation in all settings. We used different recall numbers for different methods. For the plan-retrieve method, each generation step recalls the two most relevant choices. For the task-retrieve method, we retrieve the 20 most relevant candidate steps from the task name. Stepwise Selecting and DFS methods are similar in that we both perform recalls of size 5. In the Rewrite method, we will select at most the first three steps that have not been replaced in each round, and dynamically control the recall number of each step to around 10. In all settings, we will first perform deduplication on the recall steps and then hand it over to LLM for other operations.

In addition, since the Stepwise Selecting, DFS, and Rewrite methods will iterate multiple times, we set an upper limit of 20, 30, and 20 iterations for them. These upper bounds are usually sufficient to complete the task, but if the LLM reaches the upper limit of the number of iterations, it means that the generated steps may be incomplete. If the plan complies with the rules, we still think the plan is executable, but incomplete plans will have an impact on the quality of the plan.

\section{Evaluation Details}
\label{sec:appendix-evaluation}

In the evaluation set, we randomly selected a total of 200 cases and conducted human evaluations on the three models and five methods we used in our experiments. The Spearman rank correlation coefficient between the results of human and automated is 80.76\%, which indicates that the automated evaluation results using ChatGPT present a significant consistency with those of human evaluation. Therefore, this automated assessment approach is deemed both reasonable and feasible.

\section{Different Retrieved Numbers}
\label{sec:appendix-retrieved}

\begin{figure}[htbp]
    \centering
    \begin{subfigure}[b]{0.85\linewidth}
        \centering
        \includegraphics[width=\textwidth]{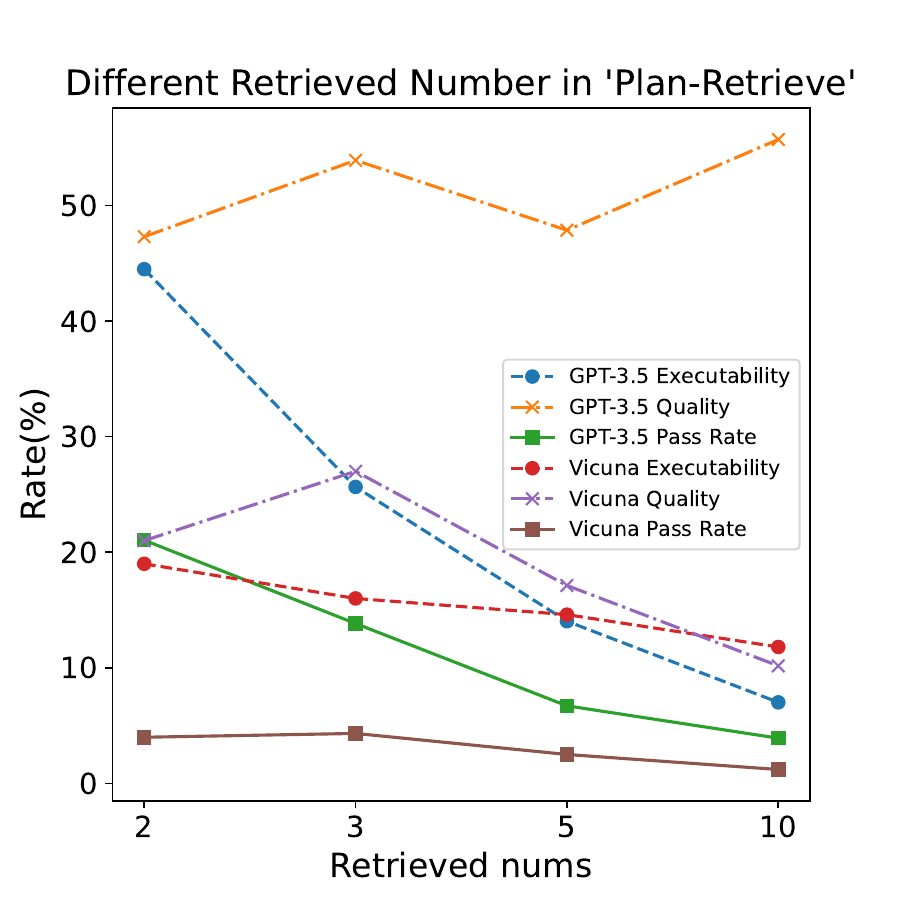}
        \caption{plan-retrieve}
        \label{fig:retrieve-plan}
    \end{subfigure}
    \hfill
    \begin{subfigure}[b]{0.85\linewidth}
        \centering
        \includegraphics[width=\textwidth]{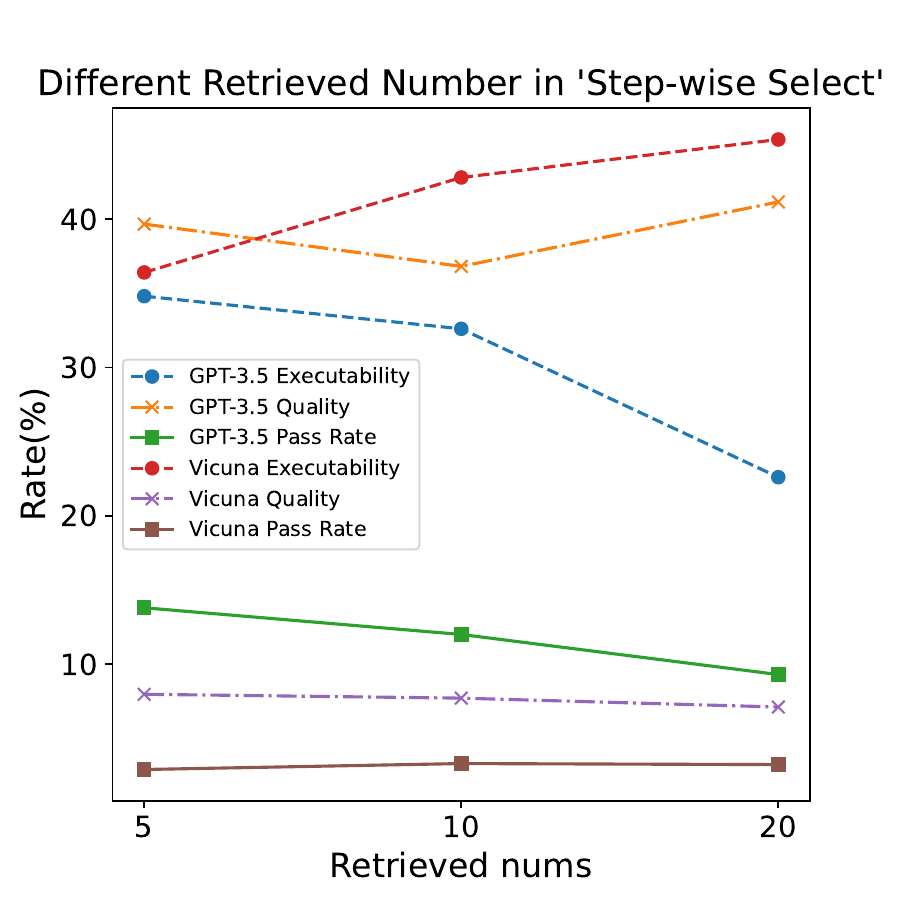}
        \caption{select}
        \label{fig:retrieve-select}
    \end{subfigure}
    \caption{Different Retrieved numbers in plan-retrieve \& select}
    \label{fig:test}
\end{figure}

The experimental results concerning the impact of the retrieved item number are illustrated in Figure 4 for both the Plan-Retrieve and Step-wise Select methods. For the plan-retrieve method, as the number of retrieved items increases, the available actions for the model to select also increase, leading decrease in plan executability, possibly due to excessive choices causing interference for the model, leading to the generation of illusory steps and consequently preventing the generated plan from being fully grounded in actions from the candidate sets. However, the generated quality shows a trend of decline after improvement.

For the select method, vicuna and GPT-3.5 show different properties. As the number of options available for vicuna increases, its executability rate will also increase. This has caused its Pass Rate to also show a slightly upward trend. The results of GPT-3.5 show a downward trend.

\section{More Result of OOD Datasets}
\label{sec:appendix-ood}
\begin{table}
\resizebox{\linewidth}{!}{
\begin{tabular}{lccc}
\toprule
 & \multicolumn{3}{c}{APIBank} \\
 \cmidrule(r){2-4}
Method & \multicolumn{1}{c}{Executability(\%)} & \multicolumn{1}{c}{Quality(\%)} & \textbf{Pass Rate(\%)} \\ \midrule
\rowcolor[HTML]{EFEFEF}
\multicolumn{4}{c}{\cellcolor[HTML]{EFEFEF}Vicuna-7B-v1.5-16k} \\ \midrule
Task-Retrieve & {\color{gray} 95.06 } & {\color{gray} 38.40 } & 36.50 \\
Plan-Retrieve & {\color{gray} 62.74 } & {\color{gray} 41.52 } & 26.05 \\
Step-wise Select & {\color{gray} 79.47 } & {\color{gray} 26.08 } & 20.72\\
DFS & {\color{gray} 100.0 } & {\color{gray} 26.81 } & 26.81 \\
Retrieve and Rewrite & {\color{gray} 91.63 } & {\color{gray} 49.59 } & 45.44 \\ \midrule
\rowcolor[HTML]{EFEFEF}
\multicolumn{4}{c}{\cellcolor[HTML]{EFEFEF}GPT-3.5}  \\ \midrule
% Task-Retrieve & {\color{gray} 73.38 } & {\color{gray} 9.84 } & 7.22 \\
Task-Retrieve & {\color{gray} 66.16 } & {\color{gray} 56.32 } & 37.26 \\
Plan-Retrieve & {\color{gray} 49.43 } & {\color{gray} 61.92 } & 30.61 \\
Step-wise Select & {\color{gray} 74.52} & {\color{gray} 43.37 } & 32.32 \\
DFS & {\color{gray} 100.00 } & {\color{gray} 35.55 } & 35.55 \\
Retrieve and Rewrite & {\color{gray} 92.40 } & {\color{gray} 47.33 } & 43.73 \\ \midrule
\rowcolor[HTML]{EFEFEF}
\multicolumn{4}{c}{\cellcolor[HTML]{EFEFEF}LLaMA-2-7B(SFT)} \\ \midrule
Task-Retrieve & {\color{gray} 96.58 } & {\color{gray} 60.83 } & \textbf{58.75} \\
Plan-Retrieve & {\color{gray} 80.61 } & {\color{gray} 59.20 } & {\ul 47.72} \\
Step-wise Select & {\color{gray} 87.83 } & {\color{gray} 41.14 } & 36.12 \\
DFS & {\color{gray} 43.02 } & {\color{gray} 26.49 } & 11.40 \\
Retrieve and Rewrite & {\color{gray} 74.43 }&{\color{gray} 61.03 } & 45.42 \\ \bottomrule
\end{tabular}
}
\caption{Detailed performance on APIBank}
\label{tab:appendix_apibank}
\end{table}

\begin{table}
\resizebox{\linewidth}{!}{
\begin{tabular}{lccc}
\toprule
 & \multicolumn{3}{c}{GPT4Tools} \\
 % & \multicolumn{3}{c}{APIBank} & \multicolumn{3}{c}{GPT4Tools} & \multicolumn{3}{c}{ToolAlpaca} & \multicolumn{3}{c}{VirtualHome} & \multicolumn{3}{c}{SayCan} \\
 \cmidrule(r){2-4}
Method & \multicolumn{1}{c}{Executability(\%)} & \multicolumn{1}{c}{Quality(\%)} & \textbf{Pass Rate(\%)} \\ \midrule
\rowcolor[HTML]{EFEFEF}
\multicolumn{4}{c}{\cellcolor[HTML]{EFEFEF}Vicuna-7B-v1.5-16k} \\ \midrule
Task-Retrieve & {\color{gray} 83.20} & {\color{gray} 19.59 } & 16.30 \\
Plan-Retrieve & {\color{gray} 54.00 } & {\color{gray} 27.96 } & 15.10 \\
Step-wise Select & {\color{gray} 73.00 } & {\color{gray} 21.64 } & 15.80 \\
DFS & {\color{gray} 99.00 } & {\color{gray} 20.71 } & 20.50 \\
Retrieve and Rewrite & {\color{gray} 59.80 } & {\color{gray} 37.29 } & 22.30 \\ \midrule
\rowcolor[HTML]{EFEFEF}
\multicolumn{4}{c}{\cellcolor[HTML]{EFEFEF}GPT-3.5} \\ \midrule
Task-Retrieve & {\color{gray} 81.40  } & {\color{gray} 28.50 } & 23.20 \\
Plan-Retrieve & {\color{gray} 67.40 } & {\color{gray} 47.92 } & 32.30 \\
Step-wise Select & {\color{gray} 76.00 } & {\color{gray} 30.92 } & 23.50 \\
DFS & {\color{gray} 100.00 } & {\color{gray} 25.00 } & 25.00 \\
Retrieve and Rewrite & {\color{gray} 90.60 } & {\color{gray} 29.36 } & 26.60 \\ \midrule
\rowcolor[HTML]{EFEFEF}
\multicolumn{4}{c}{\cellcolor[HTML]{EFEFEF}LLaMA-2-7B(SFT)} \\ \midrule
Task-Retrieve & {\color{gray} 50.80 } & {\color{gray} 52.17 } & 26.50 \\
Plan-Retrieve & {\color{gray} 61.80 } & {\color{gray} 56.63 } & {\ul 35.00} \\
Step-wise Select & {\color{gray} 17.80 } & {\color{gray} 23.60 } & 4.20 \\
DFS & {\color{gray} 4.14 } & {\color{gray} 11.60 } & 0.48\\
Retrieve and Rewrite & {\color{gray} 89.40  }&{\color{gray} 48.88 } & \textbf{43.70} \\ \bottomrule
\end{tabular}
}
\caption{Detailed performance on GPT4Tools}
\label{tab:appendix_gpt4tools}
\end{table}

\begin{table}
\resizebox{\linewidth}{!}{
\begin{tabular}{lccc}
\toprule
 & \multicolumn{3}{c}{ToolAlpaca} \\
 % & \multicolumn{3}{c}{APIBank} & \multicolumn{3}{c}{GPT4Tools} & \multicolumn{3}{c}{ToolAlpaca} & \multicolumn{3}{c}{VirtualHome} & \multicolumn{3}{c}{SayCan} \\
 \cmidrule(r){2-4}
Method & \multicolumn{1}{c}{Executability(\%)} & \multicolumn{1}{c}{Quality(\%)} & \textbf{Pass Rate(\%)} \\ \midrule
% \rowcolor[HTML]{EFEFEF}
% \multicolumn{19}{c}{\cellcolor[HTML]{EFEFEF}LLaMA-2-7B} \\ \midrule
% Task-Retrieve & {\color{gray} } & {\color{gray} } &  & {\color{gray} } & {\color{gray} } &  & {\color{gray} } & {\color{gray} } &   & {\color{gray} } & {\color{gray} } &   & {\color{gray} } & {\color{gray} } &  & {\color{gray} } & {\color{gray} } &  \\
% Plan-Retrieve & {\color{gray} } & {\color{gray} } &   & {\color{gray} } & {\color{gray} } &   & {\color{gray} } & {\color{gray} } &   & {\color{gray} } & {\color{gray} } &   & {\color{gray} } & {\color{gray} } &   & {\color{gray} } & {\color{gray} } &   \\
% Step-wise Select & {\color{gray} } & {\color{gray} } &   & {\color{gray} } & {\color{gray} } &   & {\color{gray} } & {\color{gray} } &   & {\color{gray} } & {\color{gray} } &   & {\color{gray} } & {\color{gray} } &   & {\color{gray} } & {\color{gray} } &   \\
% DFS & {\color{gray} } & {\color{gray} } &   & {\color{gray} } & {\color{gray} } &   & {\color{gray} } & {\color{gray} } &   & {\color{gray} } & {\color{gray} } &   & {\color{gray} } & {\color{gray} } &  & {\color{gray} } & {\color{gray} } &  \\
% Retrieve and Rewrite & {\color{gray} } & {\color{gray} } &  & {\color{gray} } & {\color{gray} } &  & {\color{gray} } & {\color{gray} } &  & {\color{gray} } & {\color{gray} } &  & {\color{gray} } & {\color{gray} } &  & {\color{gray} } & {\color{gray} } &  \\ \midrule
\rowcolor[HTML]{EFEFEF}
\multicolumn{4}{c}{\cellcolor[HTML]{EFEFEF}Vicuna-7B-v1.5-16k} \\ \midrule
Task-Retrieve & {\color{gray} 50.25 } & {\color{gray} 39.60 } & 19.90 \\
Plan-Retrieve & {\color{gray} 41.79 } & {\color{gray} 23.21 } & 9.70 \\
Step-wise Select & {\color{gray} 53.73 } & {\color{gray} 18.98 } & 10.20 \\
DFS & {\color{gray} 98.01 } & {\color{gray} 16.50 } & 16.17 \\
Retrieve and Rewrite & {\color{gray} 62.19 } & {\color{gray} 38.00 } & 23.63 \\ \midrule
\rowcolor[HTML]{EFEFEF}
\multicolumn{4}{c}{\cellcolor[HTML]{EFEFEF}GPT-3.5} \\ \midrule
% Task-Retrieve & {\color{gray} 47.76 } & {\color{gray} 12.50 } & 5.97 \\
Task-Retrieve & {\color{gray} 45.77 } & {\color{gray} 32.61 } & 14.93 \\
Plan-Retrieve & {\color{gray} 35.32 } & {\color{gray} 33.80 } & 11.94 \\
Step-wise Select & {\color{gray} 76.12} & {\color{gray} 21.90 } & 16.67 \\
DFS & {\color{gray} 100.00 } & {\color{gray} 19.90 } & 19.90 \\
Retrieve and Rewrite & {\color{gray} 87.06 } & {\color{gray} 33.14 } & 28.86  \\ \midrule
\rowcolor[HTML]{EFEFEF}
\multicolumn{4}{c}{\cellcolor[HTML]{EFEFEF}LLaMA-2-7B(SFT)} \\ \midrule
Task-Retrieve & {\color{gray} 99.50 } & {\color{gray} 49.25 } & \textbf{49.00} \\
Plan-Retrieve & {\color{gray} 79.10 } & {\color{gray} 45.60 } & 36.07 \\
Step-wise Select & {\color{gray} 67.66 } & {\color{gray} 32.35 } & 21.89 \\
DFS & {\color{gray} 12.62 } & {\color{gray} 22.10 } & 2.79 \\
Retrieve and Rewrite & {\color{gray} 90.55 }&{\color{gray} 49.73 } & {\ul 45.02} \\ \bottomrule
\end{tabular}
}
\caption{Detailed performance on ToolAlpaca}
\label{tab:appendix_toolalpaca}
\end{table}

\begin{table}
\resizebox{\linewidth}{!}{
\begin{tabular}{lccc}
\toprule
 & \multicolumn{3}{c}{VirtualHome} \\
 % & \multicolumn{3}{c}{APIBank} & \multicolumn{3}{c}{GPT4Tools} & \multicolumn{3}{c}{ToolAlpaca} & \multicolumn{3}{c}{VirtualHome} & \multicolumn{3}{c}{SayCan} \\
 \cmidrule(r){2-4}
Method & \multicolumn{1}{c}{Executability(\%)} & \multicolumn{1}{c}{Quality(\%)} & \textbf{Pass Rate(\%)} \\ \midrule
% \rowcolor[HTML]{EFEFEF}
% \multicolumn{19}{c}{\cellcolor[HTML]{EFEFEF}LLaMA-2-7B} \\ \midrule
% Task-Retrieve & {\color{gray} } & {\color{gray} } &  & {\color{gray} } & {\color{gray} } &  & {\color{gray} } & {\color{gray} } &   & {\color{gray} } & {\color{gray} } &   & {\color{gray} } & {\color{gray} } &  & {\color{gray} } & {\color{gray} } &  \\
% Plan-Retrieve & {\color{gray} } & {\color{gray} } &   & {\color{gray} } & {\color{gray} } &   & {\color{gray} } & {\color{gray} } &   & {\color{gray} } & {\color{gray} } &   & {\color{gray} } & {\color{gray} } &   & {\color{gray} } & {\color{gray} } &   \\
% Step-wise Select & {\color{gray} } & {\color{gray} } &   & {\color{gray} } & {\color{gray} } &   & {\color{gray} } & {\color{gray} } &   & {\color{gray} } & {\color{gray} } &   & {\color{gray} } & {\color{gray} } &   & {\color{gray} } & {\color{gray} } &   \\
% DFS & {\color{gray} } & {\color{gray} } &   & {\color{gray} } & {\color{gray} } &   & {\color{gray} } & {\color{gray} } &   & {\color{gray} } & {\color{gray} } &   & {\color{gray} } & {\color{gray} } &  & {\color{gray} } & {\color{gray} } &  \\
% Retrieve and Rewrite & {\color{gray} } & {\color{gray} } &  & {\color{gray} } & {\color{gray} } &  & {\color{gray} } & {\color{gray} } &  & {\color{gray} } & {\color{gray} } &  & {\color{gray} } & {\color{gray} } &  & {\color{gray} } & {\color{gray} } &  \\ \midrule
\rowcolor[HTML]{EFEFEF}
\multicolumn{4}{c}{\cellcolor[HTML]{EFEFEF}Vicuna-7B-v1.5-16k} \\ \midrule
Task-Retrieve & {\color{gray} 74.80 } & {\color{gray}  16.58 } & 12.40 \\
Plan-Retrieve & {\color{gray} 45.80 } & {\color{gray} 25.98 } & 11.90 \\
Step-wise Select & {\color{gray} 65.20 } & {\color{gray} 10.89 } & 7.10 \\
DFS & {\color{gray} 98.60 } & {\color{gray} 6.90 } & 6.80 \\
Retrieve and Rewrite & {\color{gray} 75.20 } & {\color{gray} 21.28 } & 16.00 \\ \midrule
\rowcolor[HTML]{EFEFEF}
\multicolumn{4}{c}{\cellcolor[HTML]{EFEFEF}GPT-3.5} \\ \midrule
Task-Retrieve & {\color{gray} 97.40} & {\color{gray} 27.21 } & 26.50 \\
Plan-Retrieve & {\color{gray} 67.00 } & {\color{gray}  56.12 } & 37.60 \\
Step-wise Select & {\color{gray} 79.00 } & {\color{gray} 38.73 } & 30.60 \\
DFS & {\color{gray} 100.00 } & {\color{gray} 32.90 } & 32.90 \\
Retrieve and Rewrite & {\color{gray} 97.00 } & {\color{gray} 48.45 } & \textbf{47.00} \\ \midrule
\rowcolor[HTML]{EFEFEF}
\multicolumn{4}{c}{\cellcolor[HTML]{EFEFEF}LLaMA-2-7B(SFT)} \\ \midrule
Task-Retrieve & {\color{gray} 99.60 } & {\color{gray} 37.65 } & 37.50 \\
Plan-Retrieve & {\color{gray} 91.80 } & {\color{gray} 46.51 } & 42.70 \\
Step-wise Select & {\color{gray} 100.00 } & {\color{gray} 34.10 } & 34.10 \\
DFS & {\color{gray} 92.40 } & {\color{gray} 38.57 } & 35.64 \\
Retrieve and Rewrite & {\color{gray} 96.20 }&{\color{gray} 48.65 } & {\ul 46.80} \\ \bottomrule
\end{tabular}
}
\caption{Detailed performance on VirtualHome}
\label{tab:appendix_virtualhome}
\end{table}

\begin{table}
\resizebox{\linewidth}{!}{
\begin{tabular}{lccc}
\toprule
 & \multicolumn{3}{c}{SayCan} \\
 % & \multicolumn{3}{c}{APIBank} & \multicolumn{3}{c}{GPT4Tools} & \multicolumn{3}{c}{ToolAlpaca} & \multicolumn{3}{c}{VirtualHome} & \multicolumn{3}{c}{SayCan} \\
 \cmidrule(r){2-4}
Method & \multicolumn{1}{c}{Executability(\%)} & \multicolumn{1}{c}{Quality(\%)} & \textbf{Pass Rate(\%)} \\ \midrule
% \rowcolor[HTML]{EFEFEF}
% \multicolumn{19}{c}{\cellcolor[HTML]{EFEFEF}LLaMA-2-7B} \\ \midrule
% Task-Retrieve & {\color{gray} } & {\color{gray} } &  & {\color{gray} } & {\color{gray} } &  & {\color{gray} } & {\color{gray} } &   & {\color{gray} } & {\color{gray} } &   & {\color{gray} } & {\color{gray} } &  & {\color{gray} } & {\color{gray} } &  \\
% Plan-Retrieve & {\color{gray} } & {\color{gray} } &   & {\color{gray} } & {\color{gray} } &   & {\color{gray} } & {\color{gray} } &   & {\color{gray} } & {\color{gray} } &   & {\color{gray} } & {\color{gray} } &   & {\color{gray} } & {\color{gray} } &   \\
% Step-wise Select & {\color{gray} } & {\color{gray} } &   & {\color{gray} } & {\color{gray} } &   & {\color{gray} } & {\color{gray} } &   & {\color{gray} } & {\color{gray} } &   & {\color{gray} } & {\color{gray} } &   & {\color{gray} } & {\color{gray} } &   \\
% DFS & {\color{gray} } & {\color{gray} } &   & {\color{gray} } & {\color{gray} } &   & {\color{gray} } & {\color{gray} } &   & {\color{gray} } & {\color{gray} } &   & {\color{gray} } & {\color{gray} } &  & {\color{gray} } & {\color{gray} } &  \\
% Retrieve and Rewrite & {\color{gray} } & {\color{gray} } &  & {\color{gray} } & {\color{gray} } &  & {\color{gray} } & {\color{gray} } &  & {\color{gray} } & {\color{gray} } &  & {\color{gray} } & {\color{gray} } &  & {\color{gray} } & {\color{gray} } &  \\ \midrule
\rowcolor[HTML]{EFEFEF}
\multicolumn{4}{c}{\cellcolor[HTML]{EFEFEF}Vicuna-7B-v1.5-16k} \\ \midrule
Task-Retrieve & {\color{gray} 70.73 } & {\color{gray} 21.12 } & 14.94 \\
Plan-Retrieve & {\color{gray} 56.71 } & {\color{gray} 21.51 } & 12.20 \\
Step-wise Select & {\color{gray} 44.51 } & {\color{gray} 4.11 } & 1.83 \\
DFS & {\color{gray} 95.12 } & {\color{gray} 2.56 } & 2.44 \\
Retrieve and Rewrite & {\color{gray} 80.49 } & {\color{gray} 15.53 } & 12.50 \\ \midrule
\rowcolor[HTML]{EFEFEF}
\multicolumn{4}{c}{\cellcolor[HTML]{EFEFEF}GPT-3.5} \\ \midrule
Task-Retrieve & {\color{gray} 91.46 } & {\color{gray}  32.00 } & 29.27 \\
Plan-Retrieve & {\color{gray} 91.46 } & {\color{gray} 49.00 } & \textbf{44.82} \\
Step-wise Select & {\color{gray} 87.80 } & {\color{gray} 30.56 } & 26.83 \\
DFS & {\color{gray} 100.00 } & {\color{gray} 11.28 } & 11.28 \\
Retrieve and Rewrite & {\color{gray} 99.39 } & {\color{gray} 41.41 } & 41.16 \\ \midrule
\rowcolor[HTML]{EFEFEF}
\multicolumn{4}{c}{\cellcolor[HTML]{EFEFEF}LLaMA-2-7B(SFT)} \\ \midrule
Task-Retrieve & {\color{gray} 98.78 } & {\color{gray} 37.96 } & 37.50 \\
Plan-Retrieve & {\color{gray} 73.17 } & {\color{gray} 42.08 } & 30.79 \\
Step-wise Select & {\color{gray} 100.00 } & {\color{gray} 31.71 } & 31.71 \\
DFS & {\color{gray} 91.50 } & {\color{gray} 16.35 } & 14.96 \\
Retrieve and Rewrite & {\color{gray} 92.68 }&{\color{gray} 46.05 } & {\ul 42.68} \\ \bottomrule
\end{tabular}
}
\caption{Detailed performance on SayCan}
\label{tab:appendix_saycan}
\end{table}

In this section, we provide more detailed performance results on the five OOD datasets. Tables~\ref{tab:appendix_apibank}, \ref{tab:appendix_gpt4tools}, and \ref{tab:appendix_toolalpaca} show the results on the Tools dataset, while Tables~\ref{tab:appendix_virtualhome} and \ref{tab:appendix_saycan} present the performance of LLMs on the Robot dataset. We can draw some similar conclusions to those found in the In-Domain dataset. For example, the Task-Retrieve method has a higher execution rate, but Plan-Retrieve achieves better plans. Additionally, the performance of methods such as Select and DFS is limited by the model's capabilities, as the model needs to follow the single-step selection instruction and choose the next step reasonably. Existing models often stop too early or too late and lack attention to previous steps, leading to poorer performance.

\section{Prompts for All Methods}
\label{sec:appendix-prompt}

Our prompt contains two parts: instruction and input. In GPT-3.5, we use instruction for system messages, while in Vicuna and SFT models, since they are not trained on different system messages, we place it at the beginning of user input.

We use 0-shot in most settings and only use 2-shot in a few such as generating initial plans and rewriting. Prompts set to 0-shot often only have formatting requirements. The rest of the operations need to match the characteristics of the dataset or are more complex, so we think they require additional information.

\begin{figure*}[htbp]
    \centering
    \includegraphics[width=\textwidth]{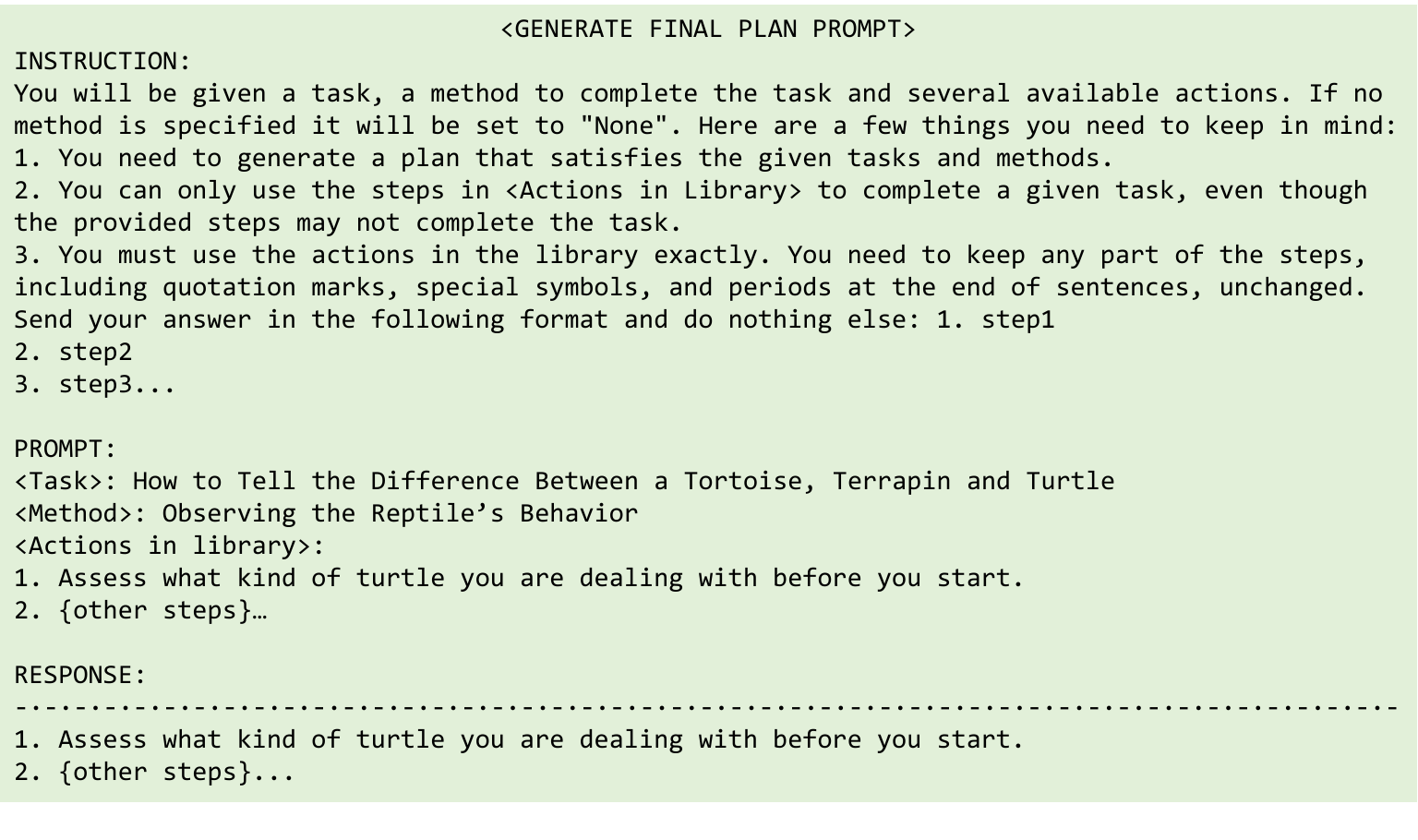}
    \caption{task-retrieve prompt.}
\end{figure*}
\begin{figure*}[htbp]
    \centering
    \includegraphics[width=\textwidth]{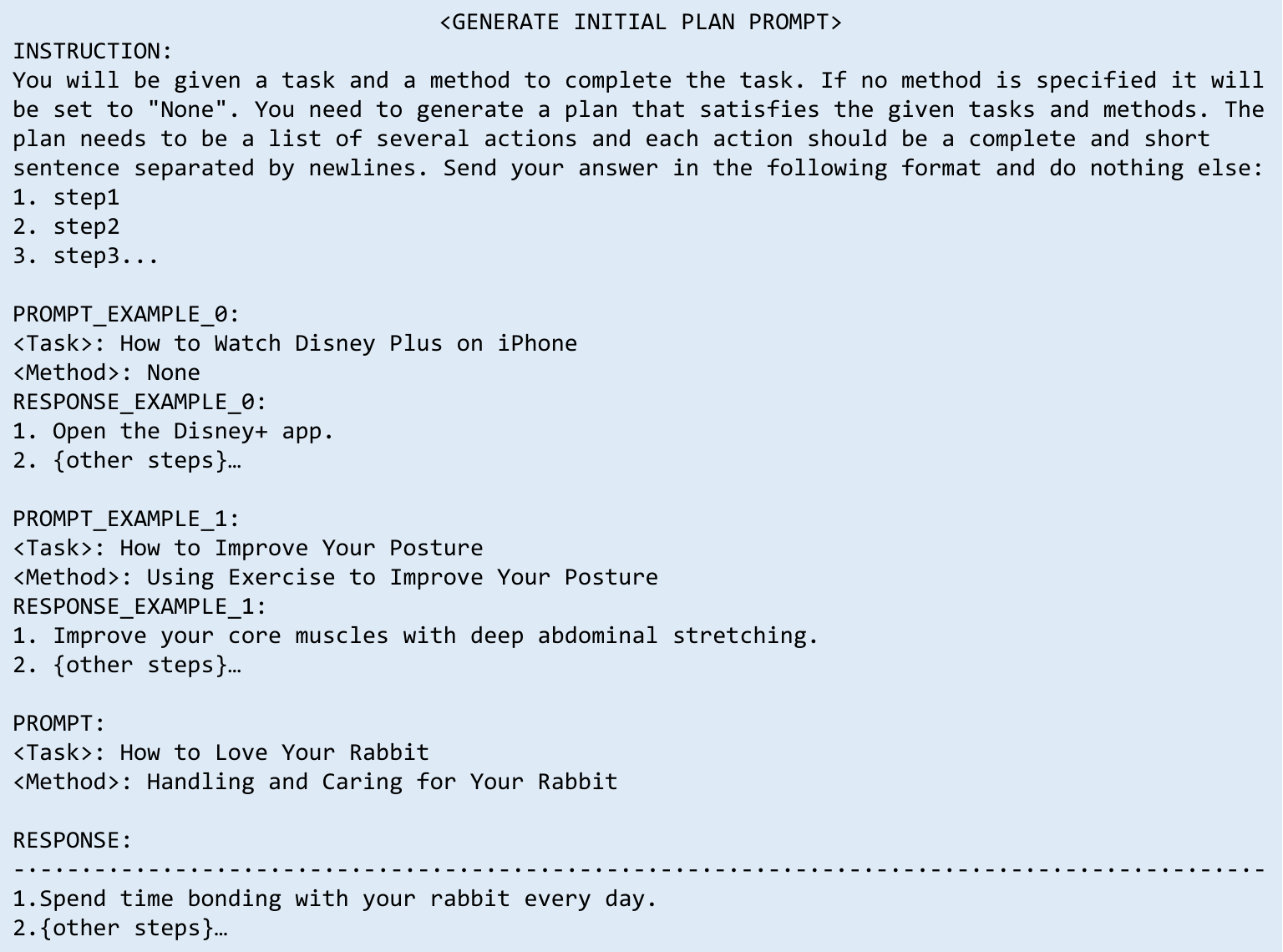}
    \caption{plan-retrieve prompt to generate initial plan.}
\end{figure*}
\begin{figure*}[htbp]
    \centering
    \includegraphics[width=\textwidth]{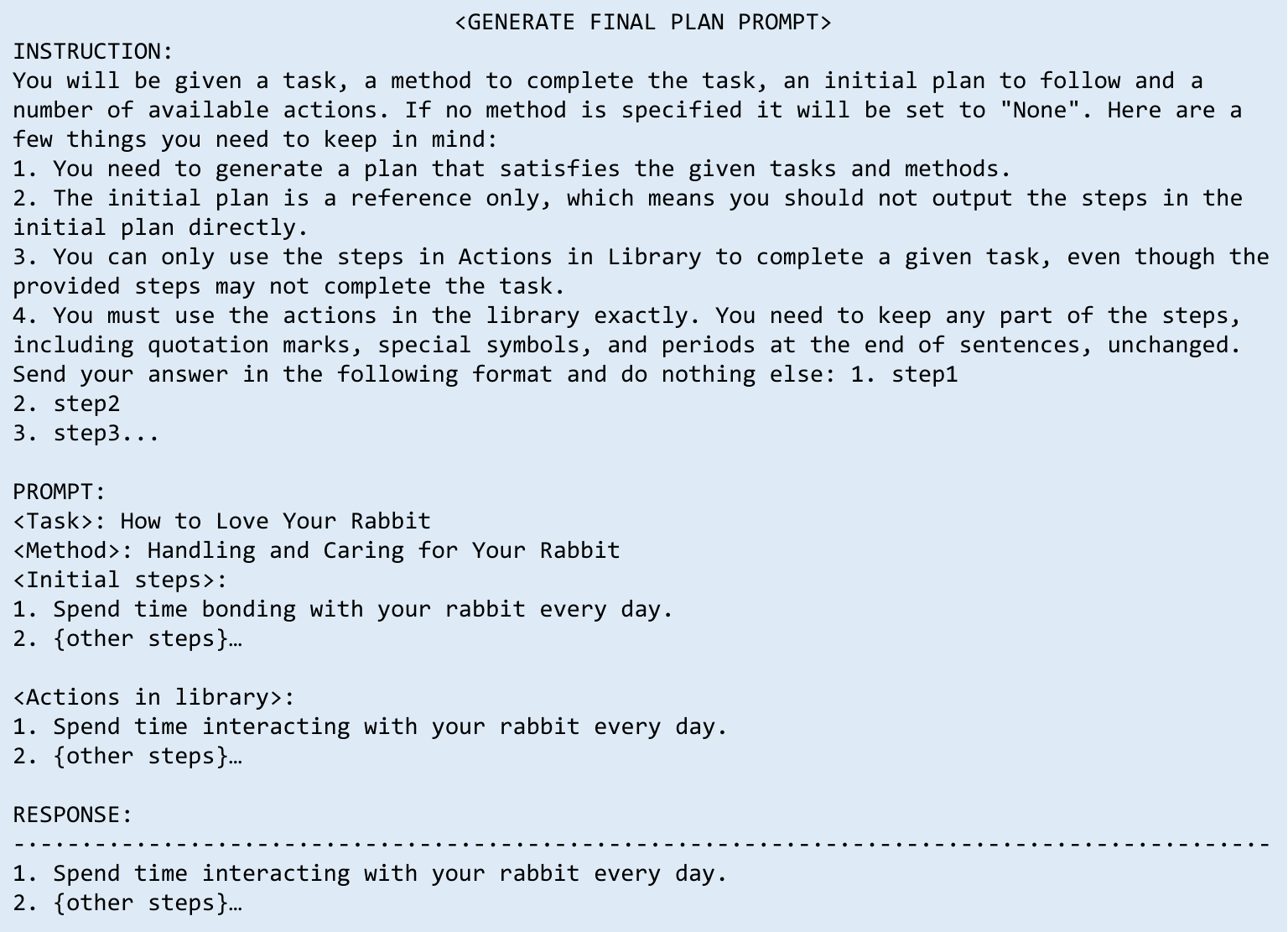}
    \caption{plan-retrieve prompt to generate final plan.}
\end{figure*}
\begin{figure*}[htbp]
    \centering
    \includegraphics[width=\textwidth]{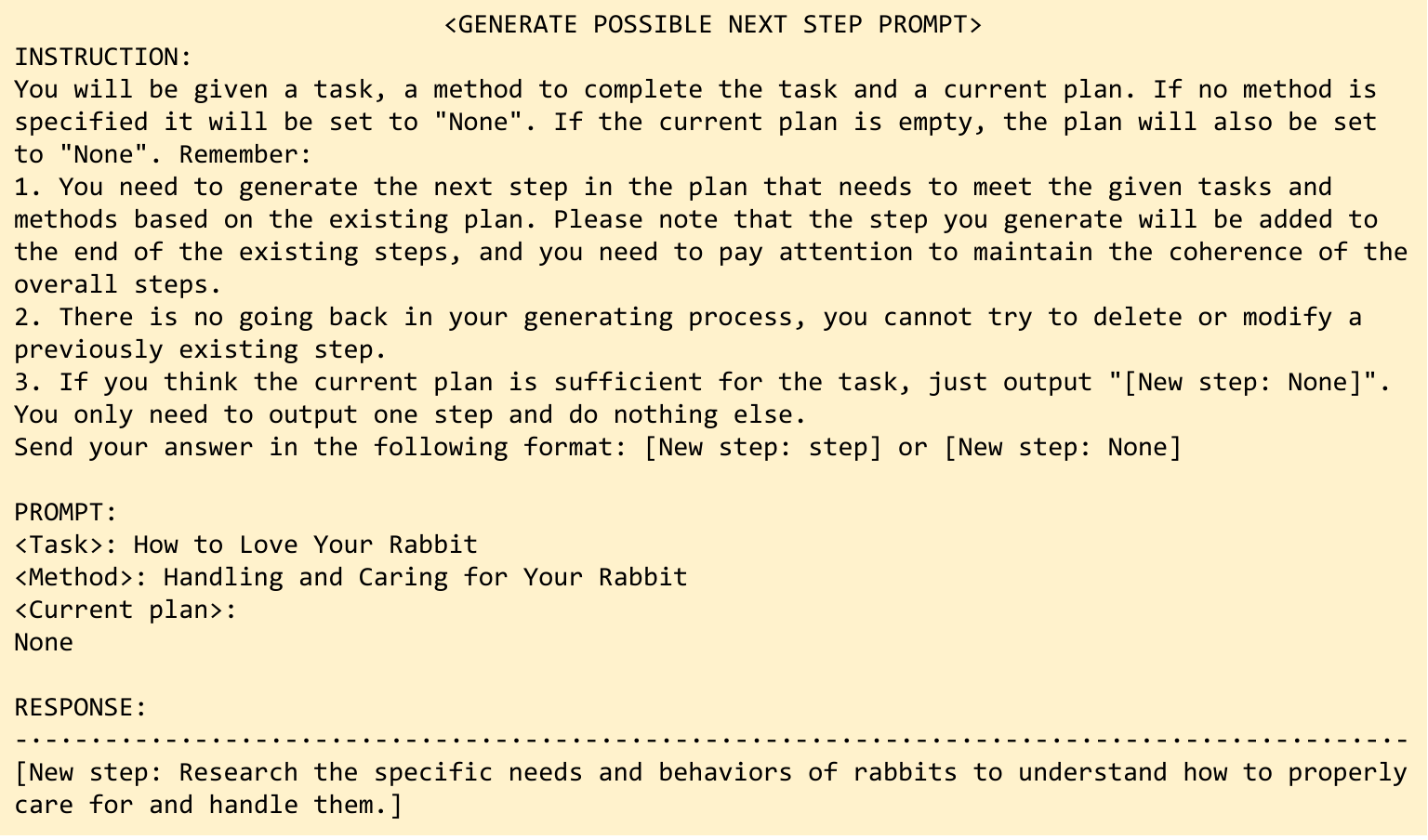}
    \caption{step-wise select prompt to generate possible next step.}
\end{figure*}
\begin{figure*}[htbp]
    \centering
    \includegraphics[width=\textwidth]{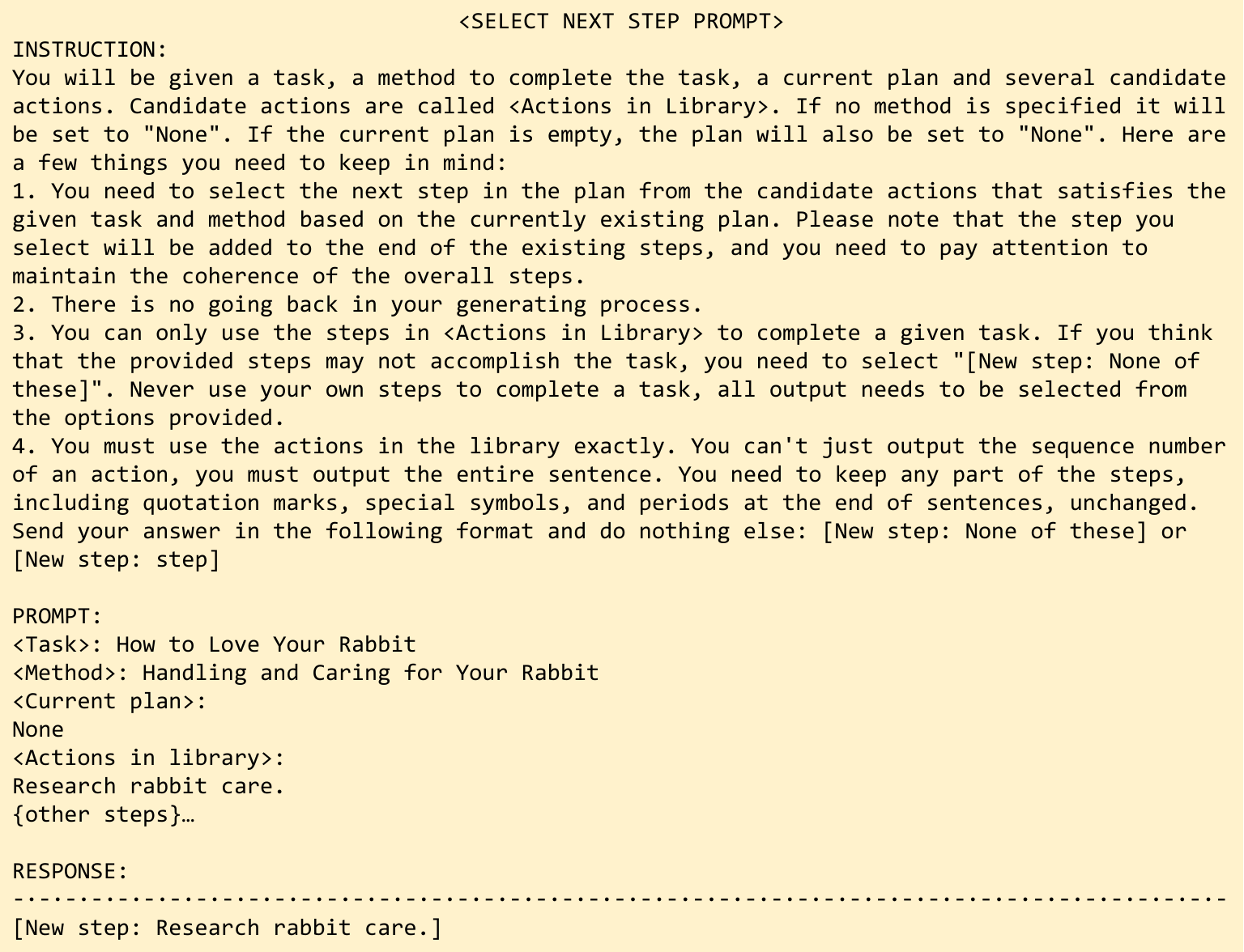}
    \caption{step-wise select prompt to select next step.}
\end{figure*}
\begin{figure*}[htbp]
    \centering
    \includegraphics[width=\textwidth]{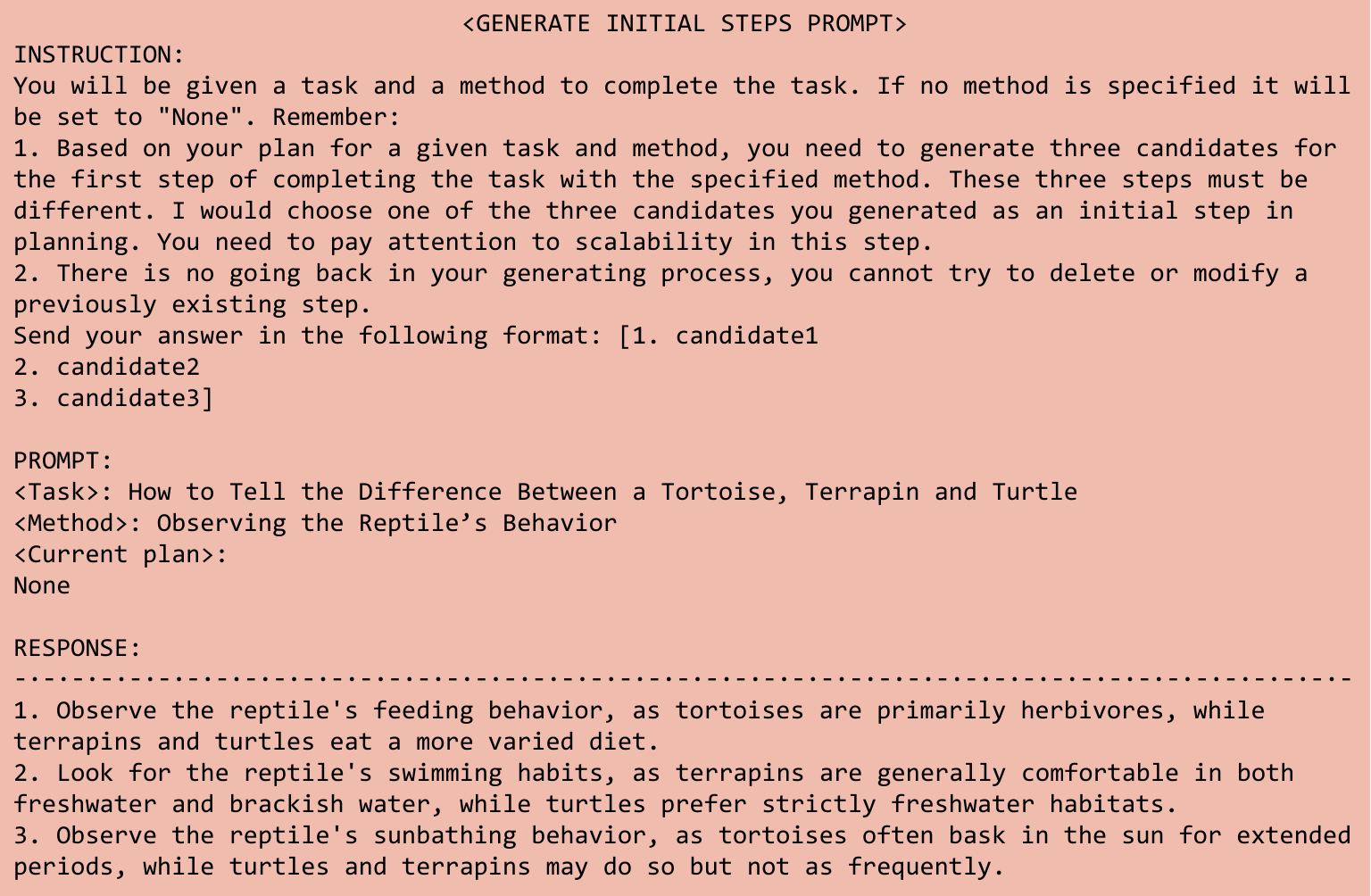}
    \caption{DFS prompt to generate initial steps.}
\end{figure*}
\begin{figure*}[htbp]
    \centering
    \includegraphics[width=\textwidth]{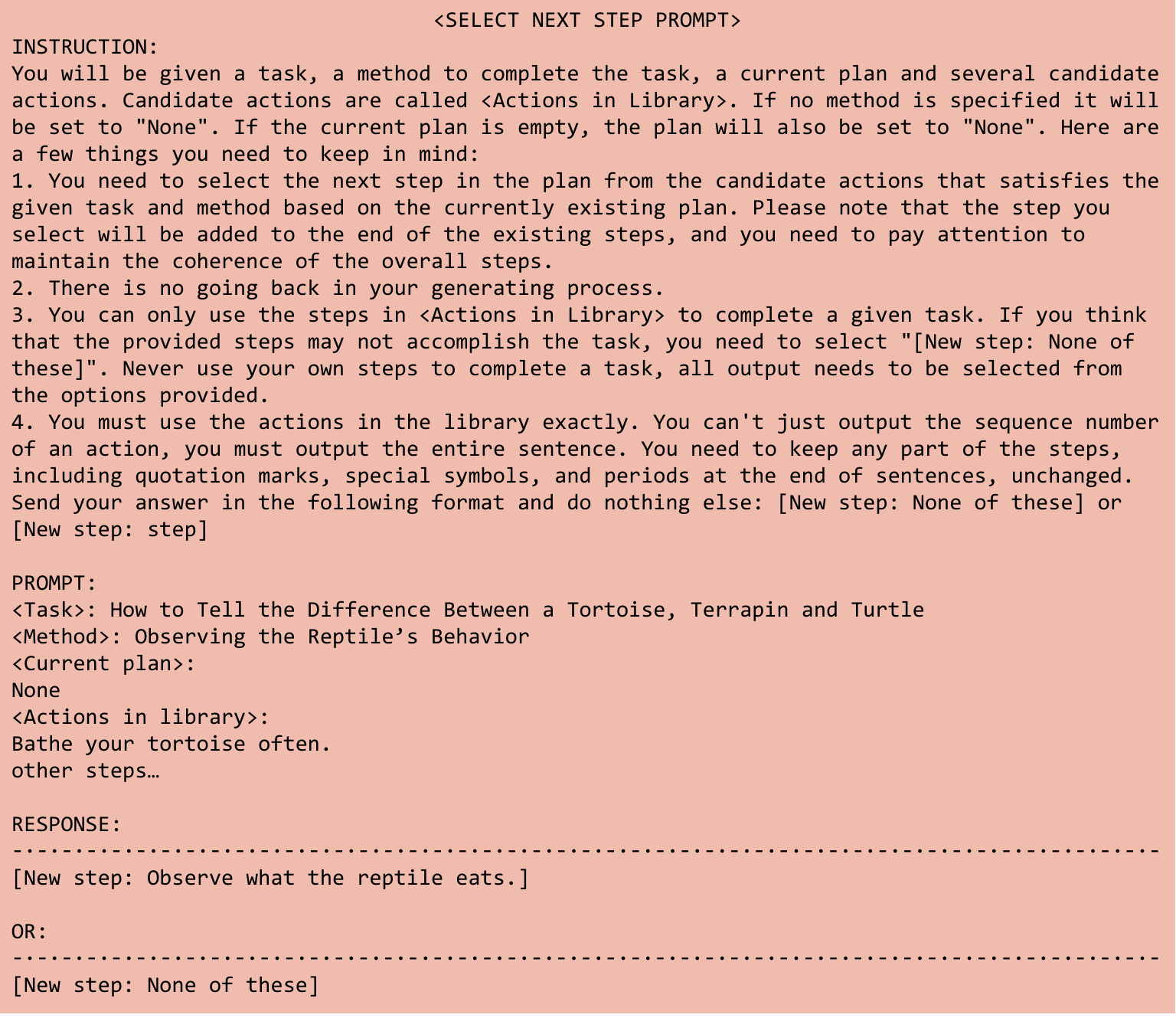}
    \caption{DFS prompt to select next step. The answer can be either new step or None.}
\end{figure*}
\begin{figure*}[htbp]
    \centering
    \includegraphics[width=\textwidth]{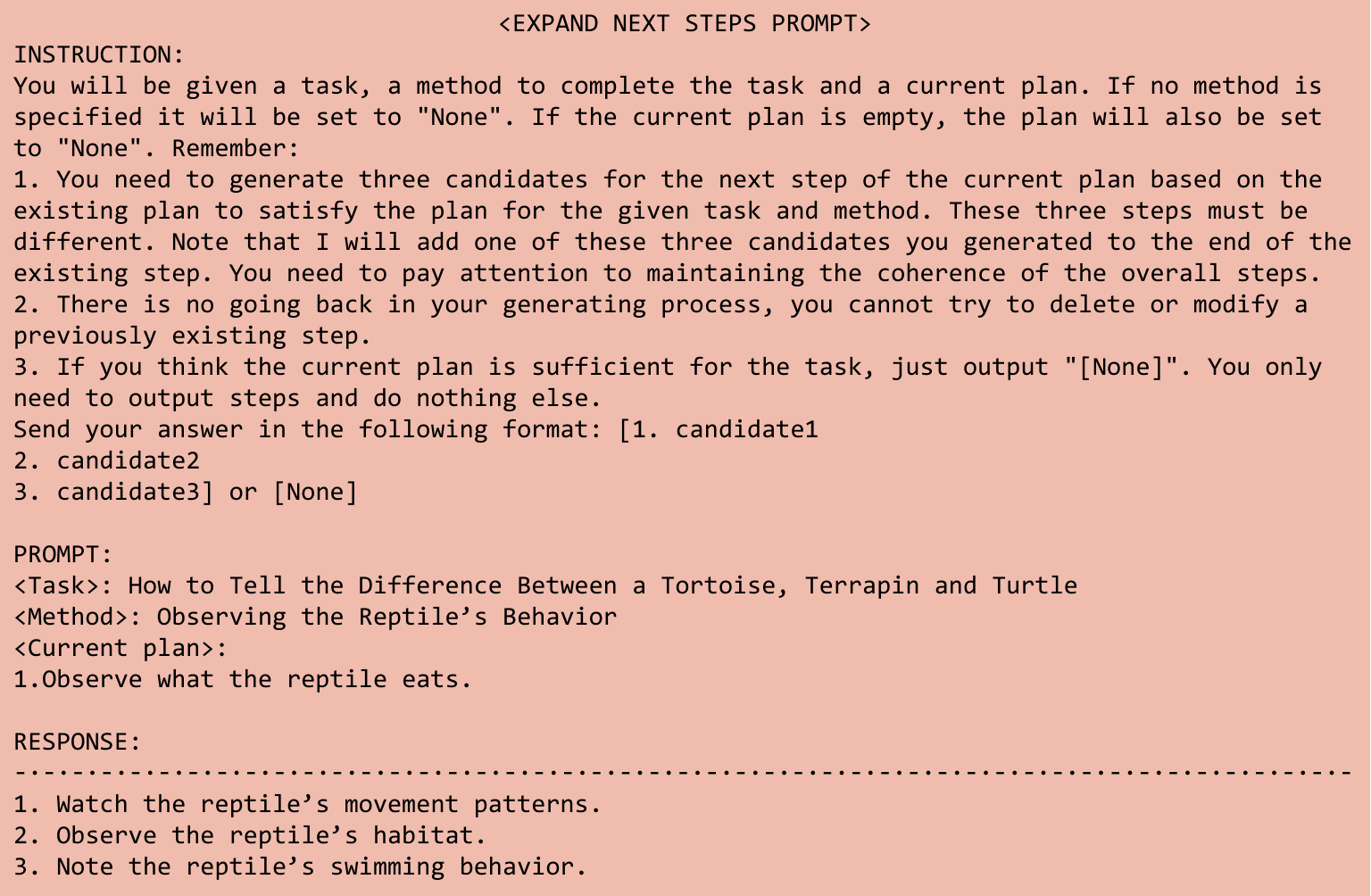}
    \caption{DFS prompt to expand nodes.}
\end{figure*}
\begin{figure*}[htbp]
    \centering
    \includegraphics[width=\textwidth]{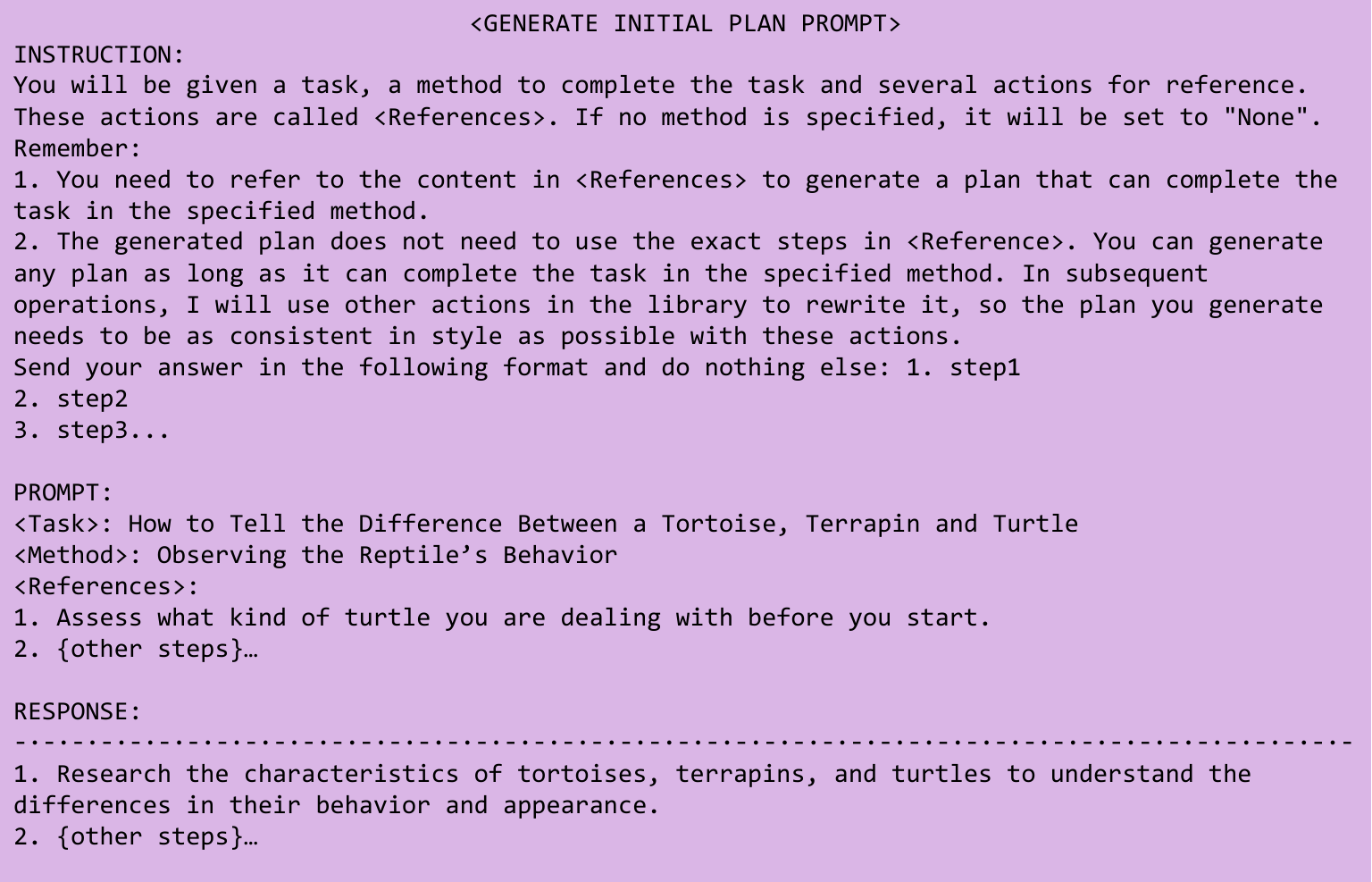}
    \caption{rewrite prompt to generate initial plan.}
\end{figure*}
\begin{figure*}[htbp]
    \centering
    \includegraphics[width=\textwidth]{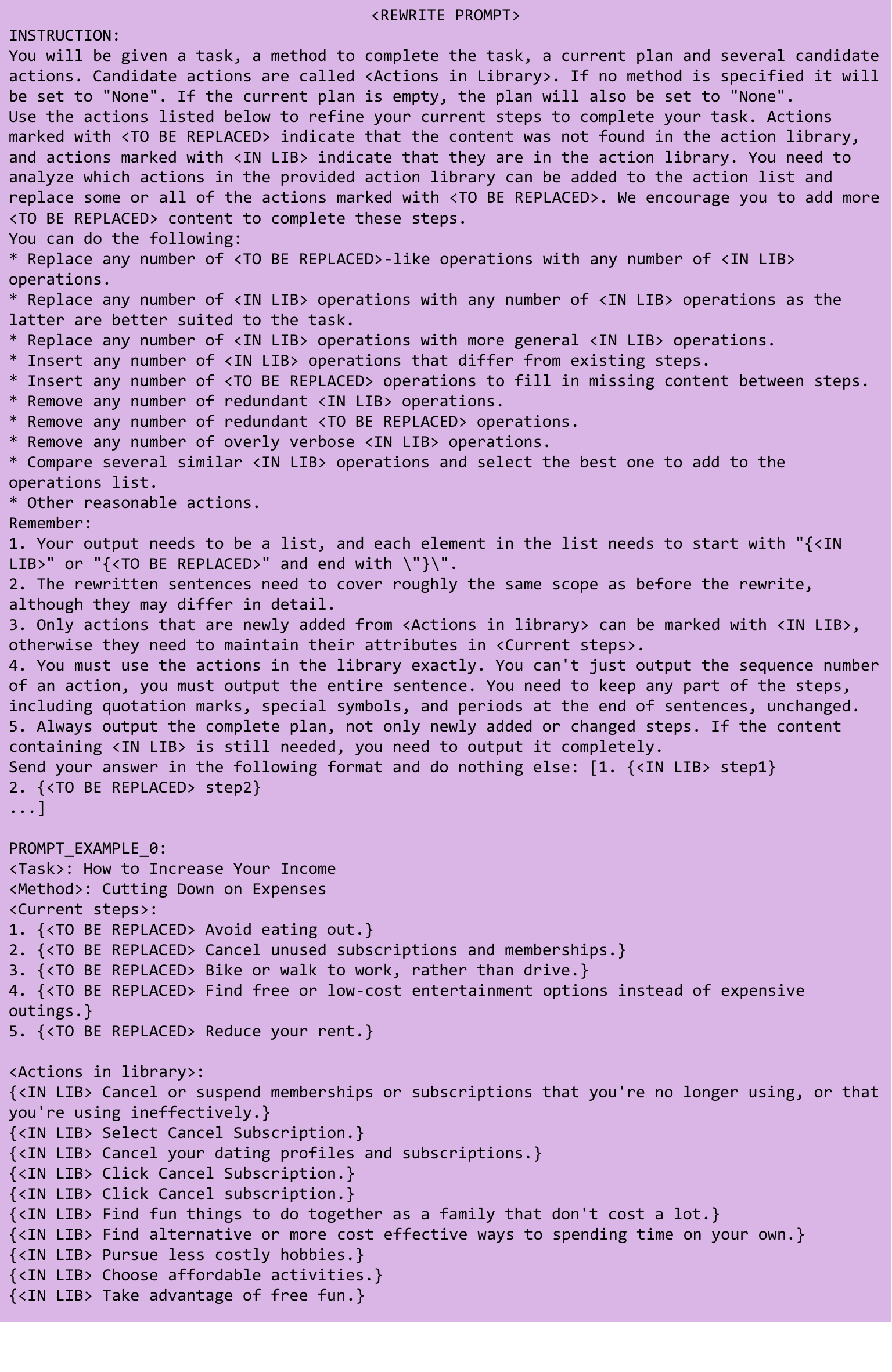}
    \caption{rewrite prompt to rewrite.}
\end{figure*}
\begin{figure*}[htbp]
    \centering
    \includegraphics[width=\textwidth]{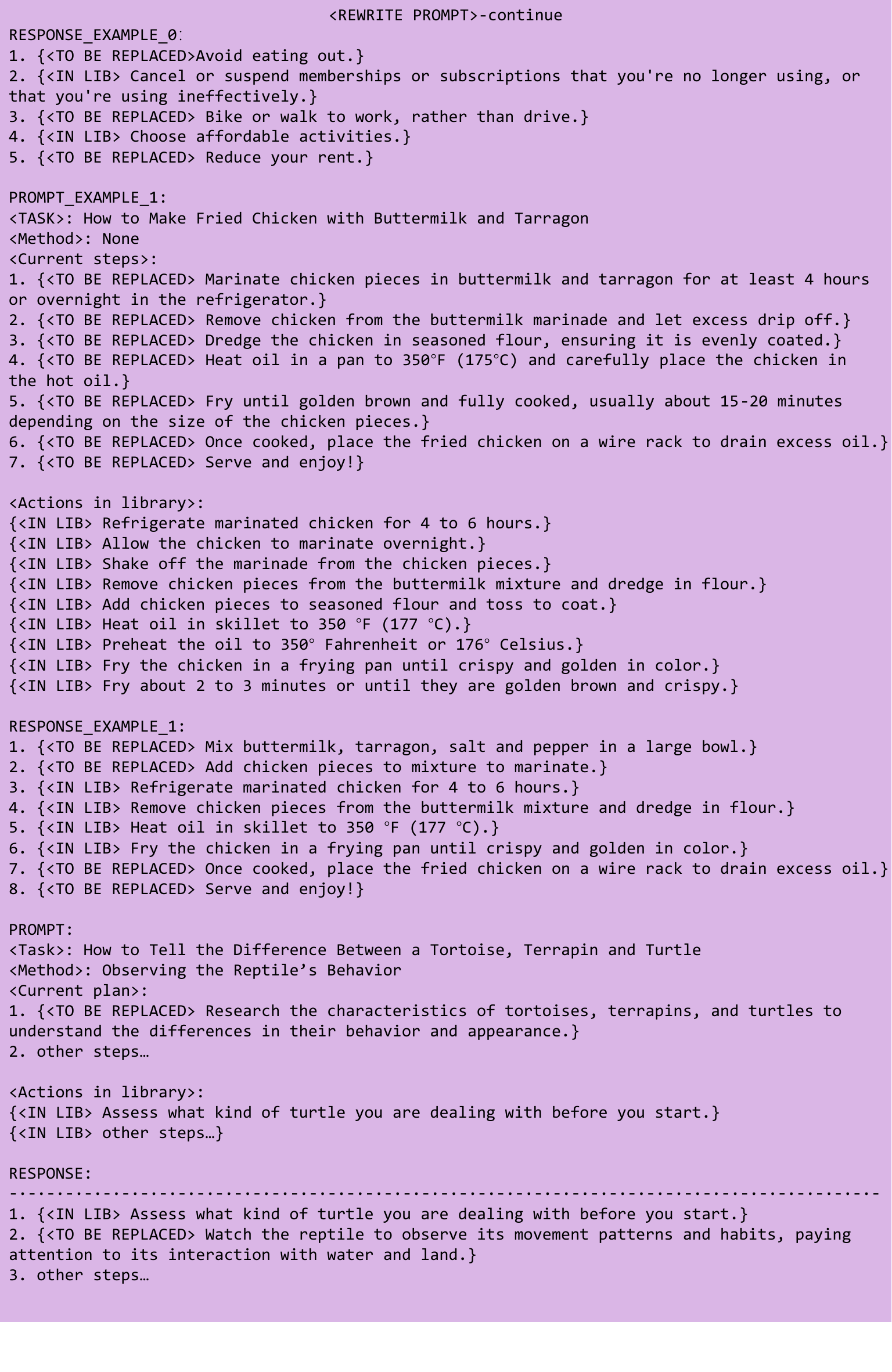}
    \caption{rewrite prompt to rewrite-continue.}
\end{figure*}
\begin{figure*}[htbp]
    \centering
    \includegraphics[width=\textwidth]{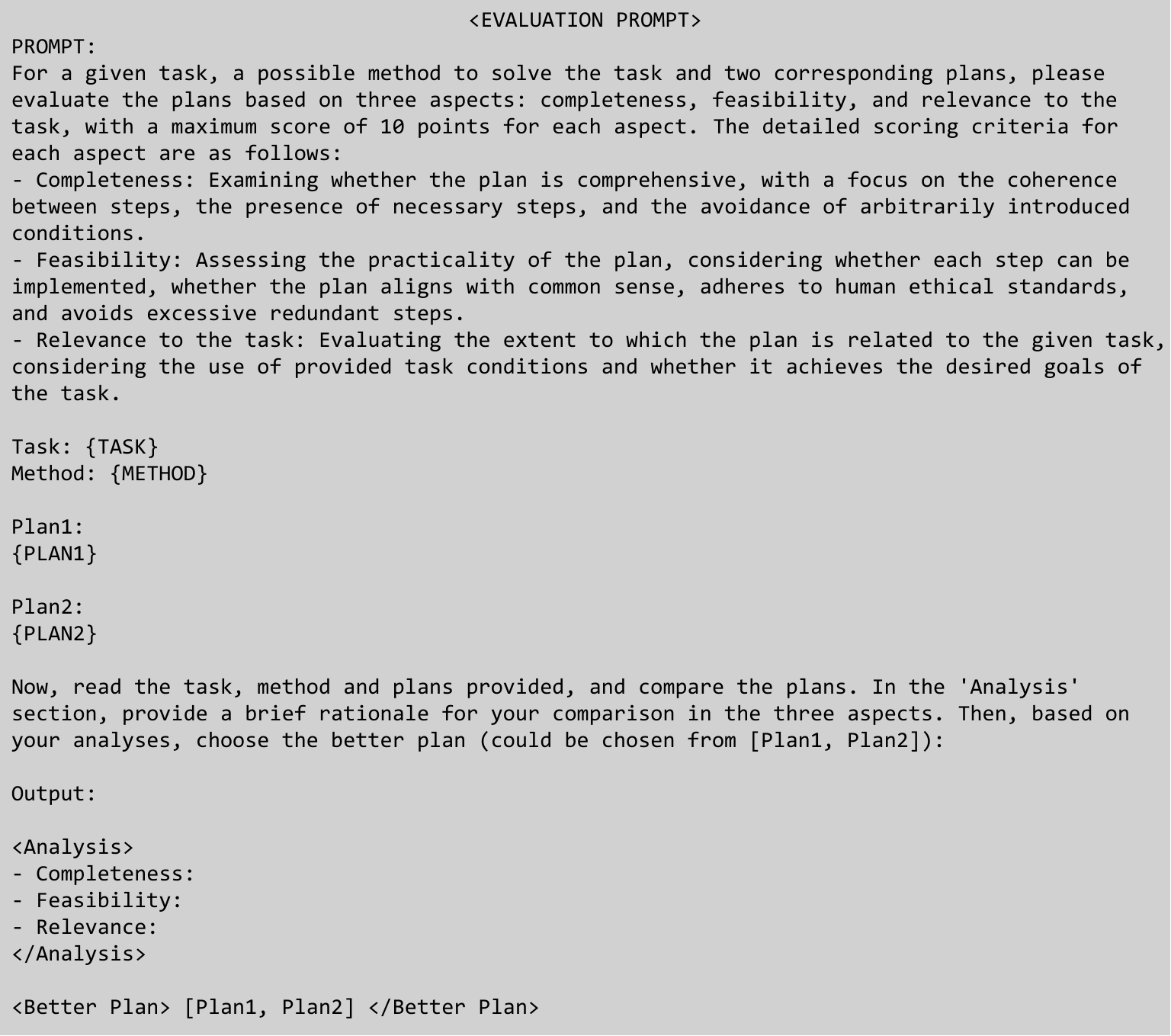}
    \caption{evaluation prompt.}
\end{figure*}

\end{document}